\documentclass[times,twocolumn,final]{elsarticle}

\usepackage{medima}
\usepackage{framed,multirow}
\usepackage{subcaption}

\usepackage{amssymb}
\usepackage{latexsym}

\usepackage{appendix}
\usepackage{amsmath}
\usepackage{url}
\usepackage{xcolor}
\usepackage{booktabs}
\usepackage{colortbl}
\usepackage{hyperref}
\usepackage[nameinlink,capitalise]{cleveref}
\usepackage{arydshln}
\definecolor{newcolor}{rgb}{.8,.349,.1}

\usepackage[flushleft]{threeparttable}
\usepackage{adjustbox}
\usepackage{soul}

\newcommand{\GG}[1]{}

\begin{document}


\begin{frontmatter}

\title{Mitosis Detection, Fast and Slow: Robust and Efficient Detection of Mitotic Figures}%

\author[1]{Mostafa  Jahanifar\corref{cor1}} 
\cortext[cor1]{Corresponding authors.}
  \ead{mostafa.jahanifar@warwick.ac.uk}
\author[1]{Adam Shephard} 
\author[1]{Neda Zamani Tajadin}

\author[1,2]{Simon Graham}
\author[1]{Shan E Ahmed Raza}
\author[1]{Fayyaz Minhas}
\author[1,2]{Nasir Rajpoot\corref{cor1}}
\ead{n.m.rajpoot@warwick.ac.uk}

\address[1]{Tissue Image Analytic (TIA) center, Department of Computer Science, University of Warwick, UK}
\address[2]{Histofy Ltd, Birmingham, UK}

\begin{abstract}
Counting of mitotic figures is a fundamental step in grading and prognostication of several cancers. However, manual mitosis counting is tedious and time-consuming. In addition, variation in the appearance of mitotic figures causes a high degree of discordance among pathologists. With advances in deep learning models, several automatic mitosis detection algorithms have been proposed but they are sensitive to {\em domain shift} often seen in histology images.
We propose a robust and efficient two-stage mitosis detection framework, which comprises mitosis candidate segmentation ({\em Detecting Fast}) and candidate refinement ({\em Detecting Slow}) stages. The proposed candidate segmentation model, termed \textit{EUNet}, is fast and accurate due to its architectural design. EUNet can precisely segment candidates at a lower resolution to considerably speed up candidate detection. Candidates are then refined using a deeper classifier network, EfficientNet-B7, in the second stage. We make sure both stages are robust against domain shift by incorporating domain generalization methods. 
We demonstrate state-of-the-art performance and generalizability of the proposed model on the three largest publicly available mitosis datasets, winning the two mitosis domain generalization challenge contests (MIDOG21 and MIDOG22). Finally, we showcase  the utility of the proposed algorithm by processing the TCGA breast cancer cohort (1,125 whole-slide images) to generate and release a repository of more than 620K mitotic figures.
\end{abstract}


\end{frontmatter}



\section{Introduction}
\begin{figure}
    \includegraphics[width=\columnwidth]{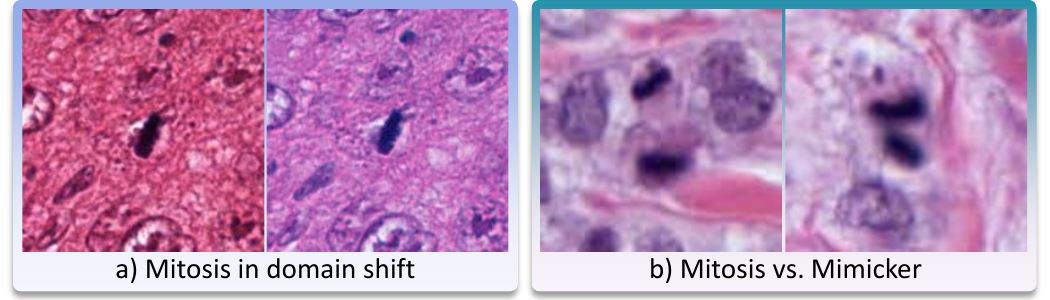}
    \caption{Two common challenges with automatic mitosis detection task.}
    \label{fig:challenge}
\end{figure}

Mitosis, a key cell-life cycle process, involves chromosome replication and separation into two nuclei, resulting in two identical cells \citep{cheeseman2008molecular}. Detection and counting of mitotic figures, particularly relevant for tumor analysis in various cancers \citep{cree2021counting}, have demonstrated a strong correlation with cell proliferation, serving as a key parameter in tumor grading systems \citep{paul2015mitosis,rakha2008prognostic}. However, the diversity in mitotic figure appearances and resemblance of imposters/mimicker cells often lead to significant inter-rater variability (see \cref{fig:challenge}.b for an example mimicker).

The rise of digital pathology (DP), driven by whole-slide scanners, has fostered the growth of Computational Pathology (CPath), which facilitates analysis of multi-gigapixel Whole-Slide Images (WSIs) \citep{graham2019hovernet,shephard2021hovernet+,koohababni2018nuclei,alemi2019nuclear}. Generally, CPath enhances objectivity and reproducibility in histopathology tasks \citep{cruz2017accurate,bizzego2019evaluating,djuric2017precision}, with Deep Learning (DL) methods providing promising avenues for automated mitotic figure detection/counting \citep{mathew2021survey,aubreville2022midog}. Nonetheless, applying machine learning to clinical practice poses challenges. Models must be robust to WSI appearance variations, stemming from differences in sample preparation, tissue types, and scanner hardware \citep{asif2021towards, aubreville2022midog}. This variability introduces \textit{domain shift} in WSIs from different scanners and sites (see \cref{fig:challenge}.a for an example of variation caused by two different scanners on the same sample).

The MItosis DOmain Generalization challenge (MIDOG21) \citep{aubreville2022midog} offered a testing ground for mitotic figure detection algorithms amidst domain shift, specifically in human breast cancer. Yet, domain shift also emerges from tissue type and species differences, affecting mitotic figure appearance \citep{bertram2019large}. Thus, robust tools for different cancer types, species, scanners, or preparation protocols are desirable. MIDOG22 \citep{aubreville2022midog22} expanded on this by considering domain shifts from different tumor types and species.

On an average WSI, among around 100,000 nuclei, 100-1,000 mitotic figures are `rare events' requiring high-resolution manual counting. Such rare event detection at high resolution ($40\times$ magnification) is taxing for humans and algorithms alike \citep{he2016resnet,lin2017retinanet,he2017mask}. To ease this, pathologists usually scout at low magnification to select a `mitotic hotspot' based on cell density and morphology, and then count mitoses at a higher magnification \citep{ellis2005pathology}. An effective mitosis detection algorithm should maintain accuracy while speeding up the counting process, especially when processing large WSIs.

In this work, we aim to resolve the accuracy and speed challenges by introducing a novel mitosis detection method,  metaphorically inspired by the `Thinking Fast and Slow' theory \citep{kahneman2011thinking} and the multi-magnification workflow used by pathologists. In particular, our approach consists of two steps: 1) segmentation of mitotic candidates (Detecting Fast) using a novel model architecture, and 2) candidate refinement by passing the candidates to a deeper classifier network to differentiate between mitotic figures and mimickers (Detecting Slow). To enable mitosis segmentation, we propose to generate mitosis masks from point annotations using NuClick \citep{koohbanani2020nuclick}. For the first time, we investigate the effect of different combinations of three well-known domain generalization techniques for mitosis segmentation. We use this knowledge to design a robust model to counter the domain shift caused by using different scanners. We show that our proposed method outperforms all other state-of-the-art (SOTA) algorithms in literature and achieved the first rank in both the MIDOG21 and MIDOG22 challenges. Furthermore, we showcase the practicality of our algorithm by detecting mitotic figures in the breast cohort of the TCGA dataset (TCGA-BRCA). In summary, the main contributions of this paper include:

\begin{enumerate}
    \item An efficient two-stage mitosis detection method based on a novel segmentation model architecture and a deep classification model. 
    \item A self-supervised training method to pretrain both encoder and decoder parts of a segmentation model.
    \item Investigation of the effect of different combinations of domain generalization techniques on the mitosis segmentation task.
    \item Release of segmentation masks for the mitotic figure in two well-known mitosis datasets (TUPAC and MIDOG) as well as the AI-generated mitosis detection dataset for TCGA-BRCA cohort which contains over 622,000 mitoses (available at \url{https://sandbox.zenodo.org/record/1227403}).
    \item Outperforming other SOTA algorithms in cross-validation experiments while being considerably faster as well as ranking 1\textsuperscript{st} in MIDOG21 and MIDOG22 challenges.
\end{enumerate}

 

\section{Related works}
\label{sec:related}

Since 2012 \citep{krizhevsky2012imagenet} convolutional neural networks (CNNs) paved the way for transformative advancements in computer vision, with impressive results in image segmentation, detection, and classification tasks \citep{li2021survey}. Their subsequent ubiquity in CPath made CNNs a cornerstone of various segmentation and classification tasks, including the detection of mitotic figures \citep{mathew2021survey,dif2020review}. The research community has responded to this phenomenon with multiple AI challenges centered around mitotic figure detection \citep{ludovic2013mitosis,veta2019tupac,aubreville2021midog21,aubreville2022midog22}. The first publicly available mitosis detection challenges and datasets were ICPR2012 \citep{ludovic2013mitosis} and its subsequent challenge MITOS-ATYPIA 2014, both of which consisted of a limited number of cases and training images. At this stage, DL-based methods were less prevalent, and the small image set was manageable. This was followed by the TUPAC16 challenge \citep{veta2019tupac}, where participants were tasked with counting mitotic figures and predicting a WSI tumor proliferation score. More recently, the MIDOG21 and MIDOG22 challenges focused on detecting mitotic figures in histology images from various scanners to address domain shift \citep{aubreville2021midog21,aubreville2022midog22}. In these contests, all participants utilized CNNs for mitotic figure detection.

\subsection{Mitosis detection}
Mitotic figure detection through DL usually involves three main approaches. The first employs patch-based classification, dividing regions of interest (ROIs) or WSIs into small patches for CNN classification. The second involves detection models that predict bounding boxes or centroid points for the mitotic figures. The third approach uses segmentation models to semantically delineate targets before determining the mitotic centroid via post-processing.

Initial DL-based methods handled this task as a classification problem. Notably, \cite{akram2018tupac} used ResNet-12 for the task on the TUPAC16 dataset \citep{veta2019tupac} and improved their model by fine-tuning it with additional mined mitotic figures. An ensemble of CNNs was proposed by \cite{tellez2018whole}, with knowledge distillation reducing computational needs and `HED stain augmentation' increasing the range of realistic H\&E stain variations for better model training. Despite their success, the severe limitation of these models lies in their inefficiency, as they need to iterate through every high-resolution WSI patch, with patches being small to contain a single mitosis.

Bounding box detection models, such as RetinaNet \citep{lin2017retinanet}, Cascade R-CNN \citep{razavi2021midog}, and EfficientDet \citep{tan2020efficientdet} are more efficient for mitosis detection than patch-based models due to their ability to process larger images, capturing more context, and enabling faster predictions. For example, both \cite{wilm2021midog} and \cite{chung2021midog} used two different versions of RetinaNet for mitosis detection while improving the domain generalization capability of their models by incorporating domain adversarial training \citep{ganin2016domain} and style transfer augmentation techniques, respectively.

Several approaches treated mitosis detection as a segmentation task using methods like Mask R-CNN \citep{he2017mask} or fully convolutional networks (FCN) such as U-Net \citep{ronneberger2015unet}, which use large image patches to reduce processing times. Notably, \cite{kausar2020smallmitosis}, \cite{fick2021midog}, and \cite{sebai2020maskmitosis} optimized Mask R-CNN for this purpose.
\cite{li2019tupac}'s FCN model, SegMitosis, used point mitosis annotations (weak labels) to form concentric circles, achieving SOTA results on the TUPAC dataset using a concentric loss in training. Additionally, \cite{yang2021midog} proposed SK-Unet, an improved U-Net model with selective kernels, that achieved the joint first rank on the MIDOG21 challenge leaderboard.
Interestingly, the top three MIDOG21 entries turned the detection task into instance-level segmentation. \cite{yang2021midog} used HoVer-Net \citep{graham2019hovernet} to generate nuclear segmentation masks and filter non-mitotic figures, creating ground truth masks. \cite{fick2021midog} manually segmented 100 mitotic figures to train an initial Mask-RCNN model, generating pixel-level segmentation.
Despite their superiority, segmentation approaches require exhaustive annotations and can be computationally costly. Our pipeline mitigates these issues by using an interactive model to generate reliable mitosis masks \citep{koohbanani2020nuclick} and performing segmentation in low resolution.


Lastly, multi-stage methods, such as the ones used by \cite{nateghi2021midog, linag2021midog, mahmood2020tupac}, have gained traction. These typically involve finding mitotic candidates using a bounding box detection model, followed by classification.
\cite{fick2021midog} implemented a two-stage process involving a Mask-RCNN for segmenting mitotic figures, followed by classification through an ensemble of DenseNet201 and ResNet50. Similarly, \cite{kondo2021domain} utilized thresholding on blue ratio images for candidate mitotic region extraction before classification with a ResNet model. While these methods improved performance, they also considerably increased computational cost.
We introduce a two-stage method with an adept segmentor for better generalization, reducing false candidates while preserving high sensitivity. By operating on downscaled images, our segmentation module considerably lowers computational costs and enhances pipeline efficiency.

\subsection{Domain generalization}

To address domain shift resulting from the varied scanner/source use, diverse approaches have been proposed, with most employing some form of color augmentation during training for enhanced algorithm generalizability amidst domain shift \citep{yang2021midog, razavi2021midog, kondo2021domain}. Techniques such as the histology-specific `HED stain augmentation'--which deconvolves an image into Hematoxylin and Eosin stain channels and perturbs them--are shown to be effective \citep{tellez2019stainaug,nateghi2021midog}. Although the effectiveness of this approach on permuting image color information during model training to cover potential stain variations has been investigated before \citep{tellez2019stainaug}, its impact on mitosis segmentation remains unexplored. Furthermore, stain normalization methods like \citep{vahadane2016stain} are widely utilized \citep{razavi2021midog,linag2021midog} to reduce the domain shift caused by sample preparation and scanner variance.


 Some methods harnessed unlabelled images from varied scanners  through image synthesis techniques, generating new image variations for training \citep{fick2021midog,chung2021midog}. Techniques like Fourier domain mixing were also employed, swapping low-frequency domain information between images for unsupervised stain normalization and augmentation, potentially increasing model generalizability \citep{yang2021midog}.


Other strategies combatting domain shift in CPath include model pretraining \citep{koohbanani2021self,vuong2022impash} and domain-adversarial training \citep{ganin2016domain, wilm2021midog, lafarge2019learning}, though their effectiveness may be task-specific \citep{stacke2020measuring}. The efficacy of these domain generalization techniques and their combinations on mitosis segmentation has yet to be explored. This study comprehensively investigates the impact of key domain generalization techniques on mitosis segmentation, aiming to identify optimal strategies.



\section{Methodology}

\begin{figure}[!t]
\centering
\includegraphics[width=\columnwidth]{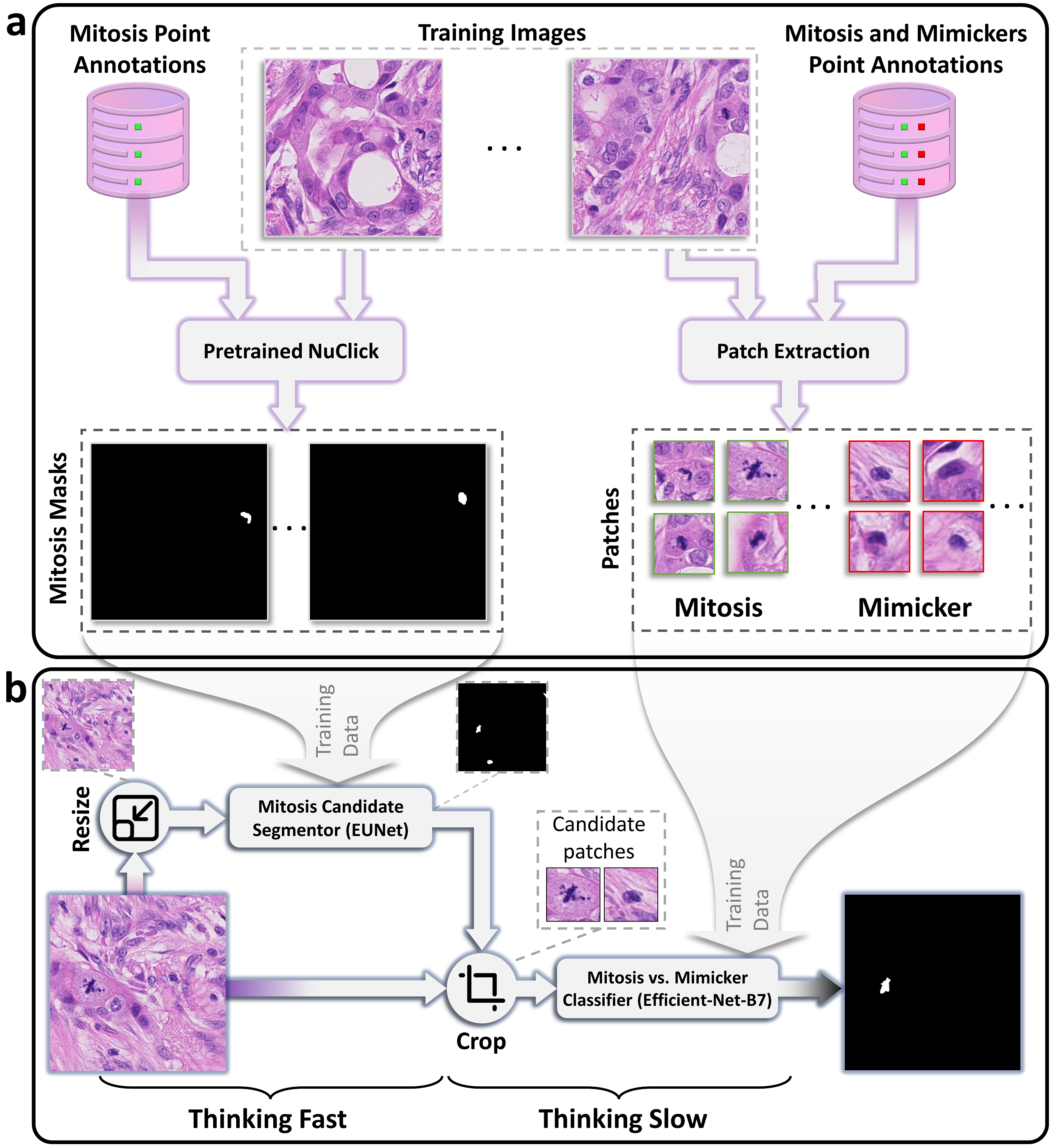}
\caption{The overview of the proposed mitosis detection method: (a) Data pre-processing steps where mitosis masks and mitosis/mimicker patches are generated, and (b) the proposed `Detecting Fast' and `Detecting Slow' systems for candidate segmentation and refinement.}
\label{fig:overview}
\end{figure}

\subsection{Overview}
The `Thinking, Fast and Slow' theory by \cite{kahneman2011thinking} considers a dichotomy between two systems of thought where system 1, or `Thinking Fast', makes decisions faster and instinctive while system 2, or `Thinking Slow', usually takes a more deliberate process to arrive at a logical conclusion. \cite{kahneman2011thinking} discusses the benefits and properties of each system and describes their importance. Many research studies in artificial intelligence have been motivated by this theory to come up with effective solutions for hard problems \citep{miech2021thinking}. 

Metaphorically inspired by this theory, we propose the `Mitosis Detection, Fast and Slow' (MDFS) framework (shown in \cref{{fig:overview}}.b) consisting of two main parts: 1) `Detecting Fast', which is responsible for finding mitosis candidates as fast and much as possible and 2) `Detecting Slow', where a deeper model refines mitosis candidates to eliminate mimickers. We approach candidate detection as a segmentation problem where mitosis masks are acquired by leveraging an interactive segmentation model, called NuClick \citep{jahanifar2019nuclick, koohbanani2020nuclick}. Because the goal of the `Detecting Fast' system is to detect plausible mitotic candidates as fast as possible with high sensitivity,  we also propose a down-sampling step in the fast system to further improve efficiency. Then, in the `Detecting Slow' system, we extract small patches around mitotic candidates at full resolution and assess them using a deeper CNN.
To make the entire framework more robust, we also include special considerations to counter the domain shift problem. We outline the proposed techniques in detail in the following sections.

\subsection{Detecting Fast: mitosis candidate segmentation}

\subsubsection{Network architecture}
\label{sec:segmodel}

We propose an efficient segmentation model architecture, called \textit{EUNet}, for the mitosis detection, which follows the encoder-decoder design of U-Net \citep{ronneberger2015unet}. Here, we use a pre-trained EfficientNet-B0 model \citep{tan2019efficientnet} as the encoder and an inverse design of that for the decoder part (using upsampling blocks instead of down-sampling). In other words, we replaced the standard convolution layers in the standard UNet architecture with the Mobile Inverted Residual  blocks coupled with a Squeeze-and-Excitation mechanism (MIRSE block).

\begin{figure}[!t]
  \centering
  \begin{minipage}[b]{1.0\columnwidth}
    \centering
    \subcaption{a) EUNet building blocks}
    \label{fig:net-a}
    \includegraphics[width=1.0\columnwidth]{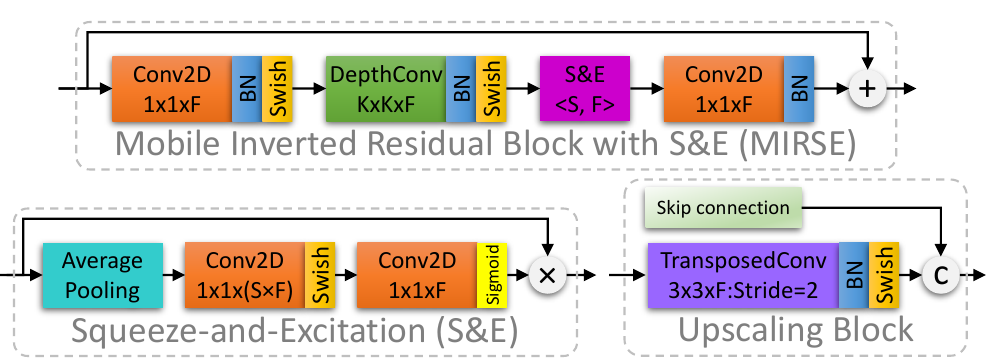}
    \qquad
  \end{minipage}
  \par
  \begin{minipage}[b]{1.0\columnwidth}
    \centering
    \subcaption{b) The arrangement of EUNet building blocks}
    \label{fig:net-b}
    \begin{adjustbox}{max width=\columnwidth}
\begin{tabular}{@{}ll@{}}
\toprule \toprule
\multicolumn{1}{c|}{Encoder}                       & \multicolumn{1}{c}{Decoder}                                                                                     \\ \hline
\multicolumn{1}{l|}{Conv2D ($K$=3, $F$=32, Stride=2)} & \begin{tabular}[c]{@{}l@{}}Upscale ($F$=320)\\ MIRSE ($K$=5, $F$=192, $R$=3)\end{tabular}                               \\ \midrule
\multicolumn{1}{l|}{MIRSE ($K$=3, $F$=16, $R$=1)}      & \begin{tabular}[c]{@{}l@{}}MIRSE ($K$=5, $F$=112, $R$=3)\end{tabular} \\ \midrule
\multicolumn{1}{l|}{MIRSE ($K$=3, $F$=24, $R$=2)\textsuperscript{*}}       & \begin{tabular}[c]{@{}l@{}}Upscale ($F$=112)\\ MIRSE ($K$=3, $F$=80, $R$=3)\end{tabular}                                \\ \midrule
\multicolumn{1}{l|}{MIRSE ($K$=5, $F$=40, $R$=2)\textsuperscript{*}}       & \begin{tabular}[c]{@{}l@{}}Upscale ($F$=80)\\ MIRSE ($K$=5, $F$=40, $R$=2)\end{tabular}                                                                  \\ \midrule
\multicolumn{1}{l|}{MIRSE ($K$=3, $F$=80, $R$=3)\textsuperscript{*}}       & \begin{tabular}[c]{@{}l@{}}Upscale ($F$=40)\\ MIRSE ($K$=3, $F$=24, $R$=2)\end{tabular}                                 \\ \midrule
\multicolumn{1}{l|}{MIRSE ($K$=5, $F$=112, $R$=3)}      & \begin{tabular}[c]{@{}l@{}}Upscale ($F$=24)\end{tabular}                                 \\ \midrule
\multicolumn{1}{l|}{MIRSE ($K$=5, $F$=192, $R$=5)\textsuperscript{*}}      & \begin{tabular}[c]{@{}l@{}}MIRSE ($K$=3, $F$=16, $R$=2)\end{tabular}                                 \\ \midrule
\multicolumn{1}{l|}{MIRSE ($K$=3, $F$=320, $R$=1)}                           & Conv2D ($K$=1, $F$=1)                                                                                               \\
\bottomrule\bottomrule
\end{tabular}
\end{adjustbox}
  \end{minipage}
  \caption{Architecture of the proposed \textbf{EUNet} for mitosis candidate segmentation. For each operation, its parameters are outlined in the parenthesis where $K$, $F$, and $R$ denote kernel size, the number of feature maps, and the number of repetitions, respectively.}
  \label{fig:net}
\end{figure}

The exact design of the MIRSE block is depicted in \cref{fig:net}a, where a sequence of a $1\times1$ 2D convolution layer, a $K \times K$ depth-wise convolution layer (to make the network lighter), a squeeze-and-excitation (S\&E) layer \citep{hu2018squeeze}, another $1\times1$ 2D convolution layer, and a residual connection (to improve back-propagation and avoid vanishing gradients) are incorporated. In all layers, the parameters $K$ and $F$ denote the kernel size and number of feature maps, respectively. It is worth noting that batch normalization (BN) \citep{he2016resnet} and Swish activation function \citep{ramachandran2017swish} are applied on the output of all convolution layers in the MIRSE block (except for the last one that only contains BN). The S\&E layers provide a self-attention mechanism inside each layer of the network to calculate the importance of different feature maps and weight them accordingly. The squeeze parameter, $S$, of the S\&E layer in this work is set to be 0.25. The `Upscaling Block' in the proposed network architecture is a $3\times3$ transposed convolution layer with a stride of 2 to increase the spatial size of input feature maps by a scale of 2. Also, this block concatenates the resulting feature maps from the same level of the encoder (retrieved via `Skip connection') with the upsampled feature maps to benefit from the high-resolution information available in the encoder part. The motivation behind incorporating elements such as Swish activations and other architectural modifications is to leverage advancements and proven benefits from related studies. These choices have been demonstrated to enhance the performance and efficiency of deep learning models in various tasks, including image classification and detection \citep{ramachandran2017swish,tan2019efficientnet,tan2020efficientdet,sandler2018mobilenetv2}.

The overall architecture design of the proposed EUNet model is described in \cref{fig:net}b where the order of different building blocks, their design parameters ($K$ and $F$), and the number of repetitions in each level ($R$) of the encoder and decoder parts are provided.
In the star-marked MIRSE blocks of the encoder path, the first convolution layer is applied with a stride of 2 to decrease the spatial resolution of feature maps by a factor of 2.

\subsubsection{Mitosis masks generation}
\label{sec:nuclick}
In order to provide stronger supervision and train our proposed segmentation model, we  use NuClick \citep{koohbanani2020nuclick, jahanifar2019nuclick} to obtain mitosis masks from mitosis point annotations. NuClick is an open-source\footnote{\url{https://github.com/mostafajahanifar/nuclick_torch}} interactive segmentation model that can generate nuclei masks from input images using point annotations as guiding signals. The outputs of NuClick have proved to be reliable in various applications \citep{koohbanani2020nuclick,shephard2021hovernet+,graham2021lizard,gamper2020pannuke}. Using NuClick, we convert all the point annotations into mitosis masks to be used during the training of our segmentation models (as shown in \cref{fig:overview}a).

\subsubsection{Training}
\label{sec:segtrain}
Our segmentation model is trained on image tiles of size $512\times512$ pixels in all experiments. However, there is a severe class imbalance when comparing patches with mitotic figures (positive patches) to those with no mitotic figures (negative patches). This may lead to a decline in model sensitivity, owing to the model seeing many more negative patches during training. To mitigate this effect, we incorporated on-the-fly under-sampling of negative patches where each batch is forced to have an equal number of positive and negative patches.


In all experiments, model training was done in two phases. In the first phase, we froze all the encoder layers, only training the decoder, for 10 epochs. For the second phase, we trained the whole network for 50 epochs. We used the Jaccard loss function \citep{jahanifar2018segmentation} and the Adam optimizer \citep{kingma2014adam} with learning rates of 0.003 and 0.0004 to optimize the model during the first and second phases, respectively.

\subsubsection{Post-processing}
\label{sec:postproc}
\begin{figure}
    \centering
    \includegraphics[width=\columnwidth]{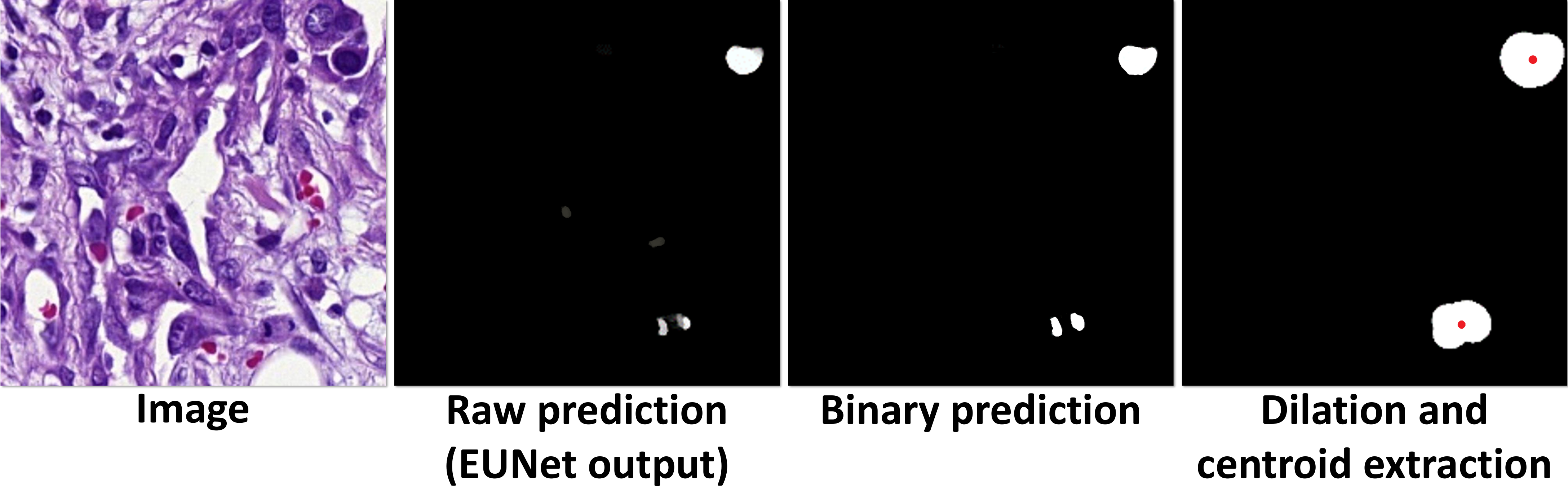}
    \caption{Mitosis segmentation post-processing steps.}
    \label{fig:postprocess}
\end{figure}

For post-processing, we performed simple thresholding of the prediction map $y_p$ to obtain a binary mitosis mask. This threshold is set based on the results from cross-validation experiments (see \ref{sec:thresh}). Then, to merge the prediction masks for mitotic figures in the anaphase or telophase, we apply a morphological dilatation operation with a disk structuring element of 18 pixels radius because daughter cells of a mitotic figure are usually closer than that radius in these phases (see \cref{fig:postprocess} for an example).
Following this, we extracted the centroids of connected components in the processed mask as mitosis candidates.

\subsection{Detecting Slow: mitosis candidate refinement}
Within the `Detecting Fast’ system of our framework, we use a mitotic candidate segmentor, operating on a down-scaled input image to improve the processing speed. Since we do not use the full-resolution input image, it is expected that the detected candidates would not be of sufficient quality. To compensate for this, we consider a deeper CNN classifier in the `Detecting Slow' system of the MDFS framework to accurately classify those candidates into mitoses or mimickers. We use EfficientNet-B7 \citep{tan2019efficientnet} as the mitotic candidate classifier which used $128\times128$ candidate patches extracted from full-resolution images as input and classify them as either mitotic figures or mimickers.

To deal with the problem of class imbalance between mitosis and mimicker categories, we incorporate under-sampling of the mimicker class as well as a weighted cross-entropy loss (where the mitosis class is given twice the weight) during the model training. These techniques allow us not to lose many real mitotic figures in the refinement phase and keep the overall sensitivity of the proposed method as high as possible.

\subsection{Addressing domain shift}
\label{sec:domain-shift}

\subsubsection{Stain normalization}
\label{sec:stainnorm}
Stain normalization is one of the mainstream techniques to address stain variation in digital pathology and various methods have been investigated to achieve it \citep{roy2018stain}. In this study, we investigate the effect of using the Vahadane stain normalization method \citep{vahadane2016stain} on mitosis segmentation. The Vahadane method has been selected as it preserves the structural properties of stained tissue samples and is robust to stain sparsity that may be found in pathology images.

In particular, \cite{vahadane2016stain} uses Sparse Non-negative Matrix Factorization (SNMF) to estimate the stain matrix $\bf{S}$ and concentration matrix $\bf{C}$ from the source and target images. Then, it scales the concentration map of the source image and combines it with the stain matrix of the target image to normalize the source image \citep{vahadane2016stain}.


\subsubsection{Stain augmentation}
We incorporate HED stain augmentation in the training of our models by randomly changing the concentration of the H\&E stains in the source image. We first used the SNMF algorithm to extract the source stain matrix $\bf{S}$ and concentrations matrix $\bf{C}$ and then we scale and shift the stain concentrations and finally convert the altered stain information back to RGB space, thus attaining an augmented image $\bf{\hat I}$:
\begin{equation}
    {\bf{\hat I}} = {\bf{I}_0}\exp \left( { - {\bf{S}}\left( {\alpha {\bf{C}} + \beta } \right)} \right)
    \label{eq:stainaug},
\end{equation}
where ${\bf{I}_0}$ is the incident intensity of the light source driven from the source image ($\bf{I}$), $\alpha\sim{U}(0.75,1.25)$ and $\beta\sim{U}( - 0.2,0.2)$ are stain concentration scale and shift factors randomly selected from uniform distributions. It is important to note that we could simultaneously perform stain normalization to a target stain matrix and stain augmentation by setting the $\bf{S}$ matrix in \cref{eq:stainaug} to a pre-extracted target stain matrix.
In this work, we use the TIAToolbox \citep{pocock2021tiatoolbox} implementation of both stain normalization and stain augmentation algorithms. Hereafter, by stain augmentation, we mean HED stain augmentation.

\subsubsection{Self-supervised learning}
\label{sec:sshl}

While labeled mitosis datasets are scarce, unlabelled Whole Slide Images (WSIs) and histology images presenting vast stain and scan quality variations are abundant. Ideally, these vast datasets can be utilized to enhance performance on limited mitosis datasets. In this pursuit, self-supervised learning (SSL) algorithms have shown success in extracting relevant visual features from unlabelled data \citep{jing2020self}. This study explores the effects of three pretraining algorithms on the mitosis segmentation task using an unlabelled histology image dataset. Until now, SSL has mainly been used in simpler CPath tasks, like patch classification \citep{stacke2020measuring,vuong2022impash,koohbanani2021self}.

\begin{figure}[!t]
\centering
\includegraphics[width=\columnwidth]{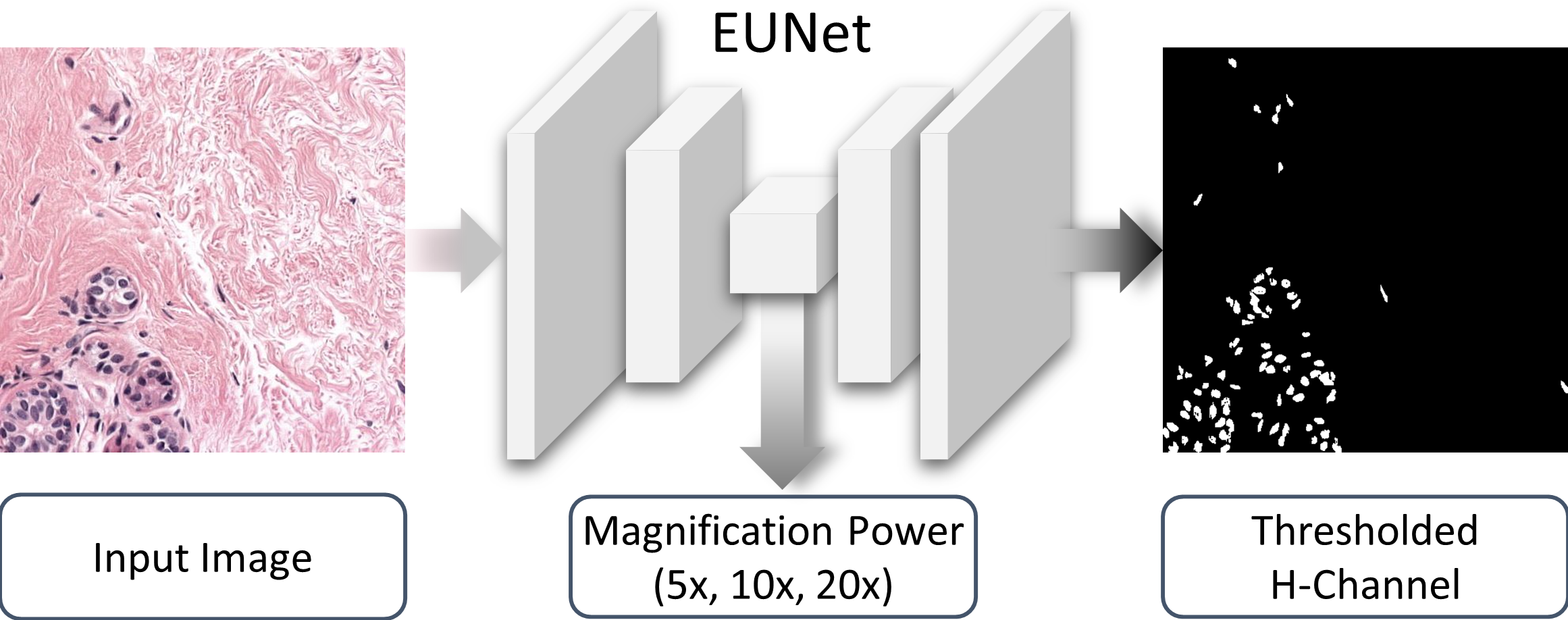}
\caption{Self-supervised Histology Learning (SSHL) for pretraining.}
\label{fig:sshl}
\end{figure}

To steer network learning towards histology-relevant features, we propose a self-supervised histology learning (SSHL) method to pretrain the entire segmentation network (both encoder and decoder). This method, inspired by \citep{koohbanani2021self}, utilizes self-supervision for two histology-related tasks: 1) image magnification power prediction and 2) Hematoxylin channel (H-Channel) segmentation as depicted in \cref{fig:sshl}. Here, two output branches are engaged, with the first branch leveraging magnification power labels for a cross-entropy loss function, and the second branch using on-the-fly generated Hematoxylin segmentation maps. These maps are obtained as follows: first, the H-channel of the input image ($\bf{H}$) is extracted using the Vahadane method, then a threshold $\tau  = 0.7 \cdot {p_{98}} + {p_2}$ is calculated for the binary conversion to $\bf{B}=\bf{H}\le\tau$ where $p_{98}$ and $p_2$ are the 98\textsuperscript{th} and 2\textsuperscript{nd} percentiles of $\bf{H}$, respectively. The binarized H-Channel, $\bf{B}$, represents the Hematoxylin-rich areas in the image. Post binarization, morphological opening operations are performed on the resultant H-Channel to eliminate spurious small objects. The training procedure is similar to the one explained in \cref{sec:segtrain}, combining the loss functions of the classification and segmentation tasks:

\begin{equation}
    {{\cal L}_{SSHL}} = {{\cal L}_{Jaccard}}\left( {\bf{B},\bf{B_p}} \right) -  \sum\limits_{i = 1}^3 {{m_i}\log \left( {{{m'}_i}} \right)}
    \label{eq:sshl},
\end{equation}
where $\bf{B_p}$, ${m'}_i$, and $m_i$ are the predicted map of the binarized H-Channel, magnification power prediction and ground truth at one of 3 categories $\{5\times, 10\times, 20\times\}$, respectively.

For comparison, we additionally pretrain the segmentation model encoder on the same dataset using a self-supervised contrastive learning (SSCL) algorithm, SimCLR \citep{chen2020simclr}, and a supervised contrastive learning method, SCL \citep{khosla2020scl}, predicting the magnification of the input image while augmenting it extensively. For all pertaining tasks, we attained more than 250,000 tiles of size $512\times512$ pixels extracted at three different levels of magnification ($5\times$, $10\times$, $20\times$) from the training set of the Camelyon16 dataset \citep{bejnordi2017diagnostic}.

\subsection{Mitosis detection in WSIs}
\label{sec:method-WSI}

The standard mitotic count or score within a $2mm^2$ hotspot Region of Interest (ROI) serves as a proxy for overall mitotic activity throughout a WSI, due to the impracticality of manual counting across an entire tissue sample \citep{rakha2008prognostic}. This approach, however, carries inherent subjectivity in ROI selection, impacting the final mitotic score. This subjectivity can be minimized by leveraging DP for mitosis detection across the entire WSI, necessitating an efficient method for accurately detecting mitotic figures within a reasonable time frame.


We thus propose a WSI processing pipeline illustrated in \cref{fig:tcga}. For each WSI, a tissue segmentation CNN is used to identify the tissue region \citep{pocock2021tiatoolbox}, from which $512\times512$ tiles with 50-pixel overlap are extracted at 0.25 microns-per-pixel resolution (approximately $40\times$ objective magnification). The proposed method then detects mitoses within these extracted patches. In this stage, the `Detecting Fast' system is trained on down-scaled images (with a scaling factor of 0.75) to identify candidates (tiles are resized accordingly). The `Detecting Slow' system refines candidates at full resolution (0.25 $\mu m$/pixel) following the MDFS method to ensure high-quality detection while minimizing processing time (refer \cref{sec:int-cval} and \cref{tab:downsample}). Once mitosis detection is completed, the mitotic hotspot region can be deterministically identified through an overlapping window search across the WSI, selecting the window with the maximum mitosis count. Subsequently, the mitotic score for the hotspot can be computed as detailed below.


Clinically, particularly in breast cancer cases, a mitotic score ($MS$) is estimated from the Mitotic Count ($MC$) in a hotspot region exhibiting high tumor activity. Specifically, the Nottingham breast cancer grading system proposes counting mitoses in a $2mm^2$ region to derive the $MS$ across three categories ($MS1, MS2, MS3$) \citep{ellis2005pathology}:

\begin{equation}
MS = \left\{ {\begin{array}{*{20}{c}}
{1,}\\
{2,}\\
{3,}
\end{array}\quad \begin{array}{*{20}{c}}
{MC \le 8}\\
{9 \le MC \le 16}\\
{MC \ge 17}
\end{array}} \right.
\label{eq:ms}
\end{equation}


\section{Materials and experiments}
\subsection{Datasets}
\label{sec:midog}
\paragraph{MIDOG}{The MIDOG21 challenge dataset \citep{aubreville2022midog}, employed for tuning parameters and model evaluation, provides a training set with 150 labeled and 50 unlabeled cases (images covering $2 mm^2$ regions), hosting 1721 mitotic figures and 2714 mimickers. Evaluation of the associated 80-case MIDOG21-test set is achieved by submitting to the challenge leaderboard, as the data is not publicly accessible. Notably, the MIDOG datasets, acquired via various scanners, exhibit domain shifts. This property informed our 3-fold \textit{leave-one-domain-out} cross-validation strategy, ensuring each fold contained images from a distinct scanner. Furthermore, we train and validate the proposed MDFS method on MIDOG22 training set \citep{aubreville2022midog22} which contains 354 labeled images from canine lung cancer, human breast cancer, canine lymphoma, human neuroendocrine tumor, and canine cutaneous cast cell tumor (3-fold cross-validation on training domains with pooled subsets for model selection). The external test set for MIDOG22 contains 100 images from different tumor types that are not provided in the training set, making mitosis detection more challenging and requiring the MDFS to capture more generalizable features. The mitosis masks for both MIDOG21 and MIDOG22 datasets are obtained using the approach explained in \cref{sec:nuclick}.}


\paragraph{TUPAC}{The TUPAC dataset \citep{veta2019tupac} offers a publicly accessible 73-case training set from three centers, hosting 1,599 mitotic point annotations. A test set of 34 cases with no available labels exists for comparison with other methodologies \citep{akram2018tupac, kausar2020smallmitosis,sebai2020maskmitosis,li2019tupac, mahmood2020tupac}. Evaluation metrics for the final test set are derived by submitting results to the TUPAC challenge organizers.}

\paragraph{ICPR2012}{ICPR2012, a widely-cited mitosis dataset, is the sole dataset providing mitosis mask annotations for its five cases \citep{ludovic2013mitosis}. Comprising 226 and 103 mitotic figures in training and test sets respectively, only the test set is used here for generalizability experiments.}

\subsection{Evaluation metrics}
\subsubsection{Detection}
Similar to the MIDOG21 challenge \citep{aubreville2022midog} and to avoid over-representation of cases with no mitoses in the evaluation metric, set-level F1 scores (F1) were used to rank the different mitosis detection methods. We also reported set-level recall/sensitivity ($Rec$) and precision ($Prc$) metrics to compare different methods more thoroughly in all cross-validation and external tests on the MIDOG dataset. However, following the TUPAC challenge convention, we use the macro-average of F1, Recall, and Precision over all images for evaluating performance on the TUPAC training and test sets \citep{veta2019tupac}.



\subsubsection{Mitotic score estimation}
\label{sec:mse}

Mitotic score ($MS$) estimation in WSIs is crucial for accurate cancer grading, particularly in breast cancer, where inaccurate $MS$ estimation can misinform treatment planning \citep{elston1991pathological}. Although other cancer types possess unique scoring systems, exploring these is beyond the study's scope. However, understanding the error in breast cancer $MS$ estimation is essential for this work.


While neither MIDOG nor TUPAC datasets provide ground truth $MS$ information, all MIDOG images and 50 TUPAC images (cases 24-73) cover approximately $2mm^2$ sample area. Using available mitotic count ($MC$) data (from GT annotations), we estimate each image's expected $MS$ via \cref{eq:ms}.


Since $MS$ estimation is an ordinal regression task, it requires an appropriate evaluation metric. Quadratic Weighted Kappa (QWK) is commonly employed for similar tasks \citep{veta2019tupac}, but data population imbalance in available datasets and real-world data render it unsuitable for this task. We, therefore, propose a mitotic score error—an average category-based mean squared error—to evaluate algorithm performance on the $MS$ estimation task:
\begin{equation}
    \overline {ME}  = \frac{1}{3}\sum\limits_{s = 1}^3 {\sum\limits_{n \in {\bf{T}}_s } {\frac{{{{(\widehat {M{S_n}} - M{S_n})}^2}}}{{{N_s}}}} },
    \label{eq:mse}
\end{equation}
where ${\widehat{M{S_n}}}$ and ${M{S_n}}$ are GT and predicted mitotic scores for case $n$, respectively. ${{\bf{T}}_s} = \{n|\widehat {M{S_n}} = s\}$ and $N_{s}$ are the set of all cases belonging to each mitotic score category  $s\in\{1,2,3\}$ and their respective population.
The squared error term emphasizes catastrophic prediction errors ($MS3$ predicted as $MS1$ or vice versa) and calculating the mean squared error for each category separately mitigates bias towards higher population categories.

\subsection{Validation experiments}
\subsubsection{Internal cross-validation}
\label{sec:int-cval}


The proposed MDFS method's performance was evaluated through cross-validation on the MIDOG21 and TUPAC training datasets. Results for the proposed EUNet segmentation model ('Detecting Fast' system) and the full pipeline (MDFS) are shown in \cref{tab:downsample} for the MIDOG21 dataset, with F1 scores of 0.754 and 0.785 respectively. Our method surpasses UNet \citep{ronneberger2015unet}, RetinaNet (with ResNet-50 backbone) \citep{lin2017retinanet}, and EfficientDet (with Efficient-Net-B4 backbone) \citep{tan2020efficientdet} by 18\%, 6\%, and 5\% in F1, respectively. Similar improvements are evident in both recall and precision metrics.


MDFS outperformed all SOTA mitosis detection methods when cross-validated on the TUPAC dataset, achieving a macro F1 of 0.781 (\cref{tab:tupac-cval}). This score outstrips the strongest reported TUPAC results by 9\% \citep{akram2018tupac}. Additionally, the proposed segmentation model, EUNet, demonstrated a high standalone detection accuracy.

It is important to mention that based on the results in \cref{tab:downsample}, we select the scaling factor (Scl) of 0.75 for the rest of the validation experiments reported in this study (except for the results reported in ablation studies -- \cref{sec:ablation}). Please refer to \cref{sec:time} to find more details about the added value of the proposed `Detection Fast and Slow` systems.

\begin{table*}[!ht]
\centering
\setlength{\tabcolsep}{5pt}
\caption{Results of internal cross-validation experiments on the MIDOG21 training set as well as the effect of down-scaling the image on the performance of the candidate segmentation and the full detection pipeline (Time of EUNet at segmentation scale (Scl) of 1 is the reference for Speed gain calculation). The reported time is the average ROI processing duration measured in seconds.}
\begin{tabular}{llccccc} 
\toprule
\toprule
                                         & \textbf{Scl} & \textbf{F1} & \textbf{Rec} & \textbf{Prc} & \textbf{Time} & \textbf{Speed}  \\  \hline
UNet           & 1          & 0.601$\pm$0.132          & 0.531$\pm$0.154          & 0.693$\pm$0.075       & 9.26 & 0.29$\times$   \\
RetinaNet      & 1          & 0.720$\pm$0.122          & 0.726$\pm$0.095          & 0.714$\pm$0.110       & 9.88 & 0.28$\times$   \\
EfficientDet   & 1          & 0.726$\pm$0.115          & 0.731$\pm$0.107          & 0.722$\pm$0.088   & 3.05 & 0.90$\times$ \\                            
\hline
\multirow{3}{*}{EUNet}  & 0.5            & 0.714$\pm$0.006       & 0.714$\pm$0.053        & 0.715$\pm$0.053  & \textbf{0.70}      & \textbf{3.94$\times$}   \\
                        & 0.75           & 0.740$\pm$0.013       & 0.782$\pm$0.039        & 0.704$\pm$0.057       & 1.46 & 1.88$\times$                   \\
                        & 1              & 0.754$\pm$0.033       & \textbf{0.824}$\pm$0.035        & 0.695$\pm$0.069     &2.75   & 1.00$\times$   \\ 
\hline
\multirow{3}{*}{MDFS} & 0.5            & 0.768$\pm$0.011       & 0.744$\pm$0.049        & 0.794$\pm$0.035  & 0.76      & 3.59$\times$    \\
                      & 0.75           & 0.781$\pm$0.006       & 0.764$\pm$0.025        & 0.799$\pm$0.024     & 1.53   & 1.80$\times$    \\
                      & 1              & \textbf{0.785}$\pm$0.004       & 0.771$\pm$0.020        & \textbf{0.801}$\pm$0.021     & 2.80   & 0.98$\times$    \\
\bottomrule \bottomrule
\end{tabular}
\label{tab:downsample}
\end{table*}

 \begin{table*}[!ht]
    \centering
    \caption{Results of internal cross-validation experiments on the TUPAC training set. Reported metrics are macro averages following the TUPAC convention.}
    \begin{threeparttable}
    \begin{tabular}{@{}lccc@{}}
    \toprule\toprule
    \multicolumn{1}{c}{Method}                                            & \multicolumn{1}{c}{F1*} & \multicolumn{1}{c}{Rec*} & \multicolumn{1}{c}{Prc*} \\ \hline
    SmallMitosis   \citep{kausar2020smallmitosis}         & 0.599                  & 0.873                   & 0.456                   \\
    EfficientDet   \citep{tan2020efficientdet}            & 0.608                  & 0.682                   & 0.549                   \\
    \cite{mahmood2020tupac} & 0.641                  & 0.642                   & 0.641                   \\
    MaskMitosis  \citep{sebai2020maskmitosis}             & 0.660                  & 0.689                   & 0.633                   \\
    \cite{akram2018tupac}     & 0.690                  & 0.661                   & 0.722                   \\
    SegMitos \citep{li2019tupac}                          & 0.717                  & -                       & -                       \\ \midrule
    EUNet                                                 & 0.771$\pm$0.062                  & \textbf{0.874}$\pm$0.081                  & 0.786$\pm$0.054                   \\
    MDFS (proposed)                                                       & \textbf{0.781}$\pm$0.016                  & 0.834$\pm$0.019                   & \textbf{0.851}$\pm$0.030                  \\ \bottomrule\bottomrule
    \end{tabular}
\end{threeparttable}
\label{tab:tupac-cval}
\end{table*}

\begin{figure*}[!h]
    \centering
    \includegraphics[width=0.85\textwidth]{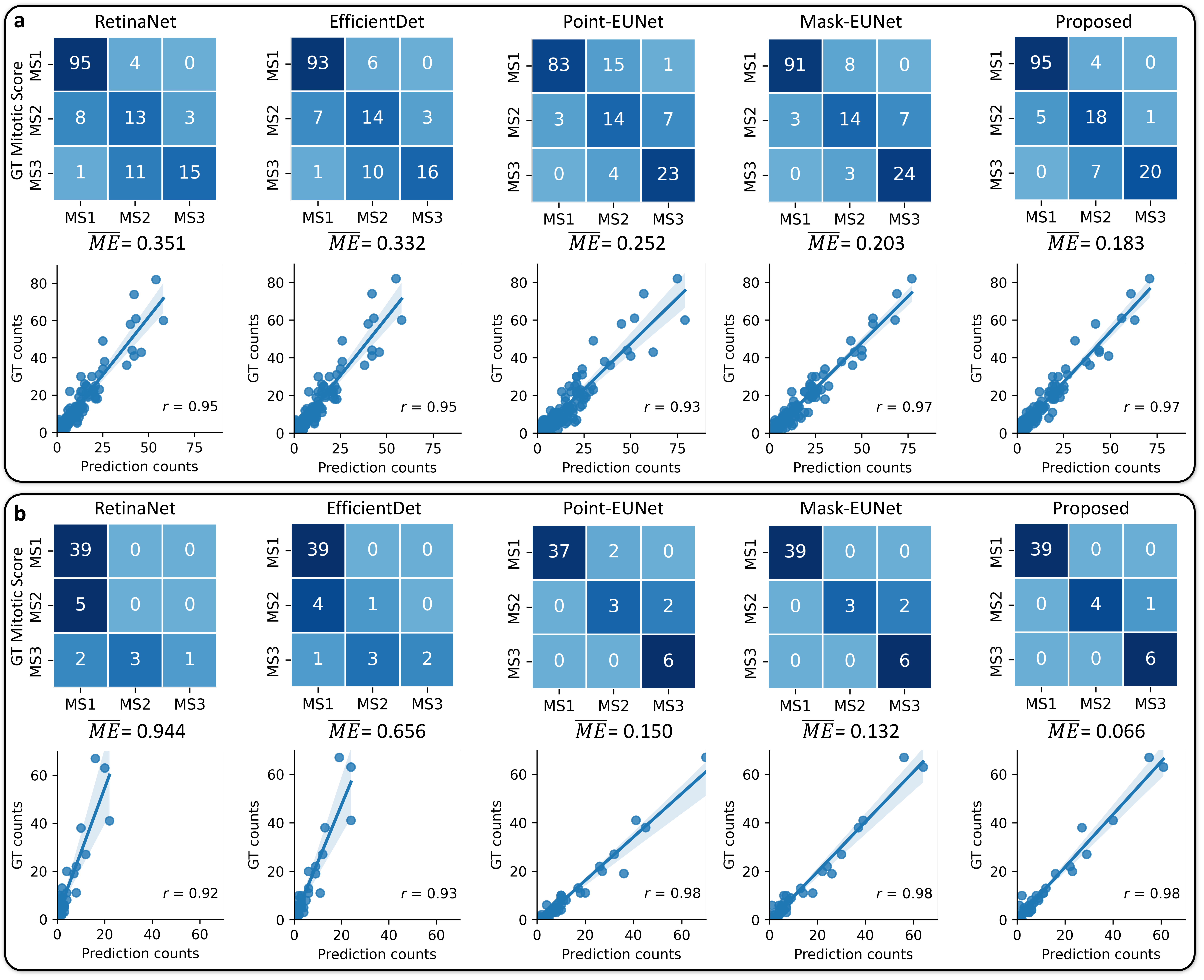}
    \caption{Confusion matrices for mitosis-score estimations (MS1, MS2, MS3) and mitosis score error ($\overline{ME}$), with correlation plots and Pearson's correlation coefficient, are reported for mitosis-count predictions from five different methods when evaluated on the MIDOG21 (a) and TUPAC (b) datasets.}
    \label{fig:mse}
\end{figure*}

For mitotic score estimation, we assessed five algorithms on the MIDOG21 and TUPAC datasets. Confusion matrices and correlation plots for the mitotic count and mitotic score estimations are shown in \cref{fig:mse}, along with the proposed mitotic score error $\overline {ME}$ and Pearson’s correlation coefficient ($r$). The MDFS method produced the lowest $\overline {ME}$ values of 0.183 and 0.066 for the MIDOG21 and TUPAC datasets, respectively, significantly outperforming the baseline (RetinaNet). Our method's predicted mitosis counts also exhibited a strong correlation with GT counts, with Pearson’s $r$ values of 0.97 and 0.98 for the MIDOG21 and TUPAC datasets, respectively.

The data in \cref{fig:mse} shed light on ROI scoring performance. While baseline RetinaNet and our method achieve F1 scores of 0.720 and 0.785 respectively on the MIDOG21 dataset, the difference in correlation scores is only 2\%—the standard metric for ROI-based mitosis assessment \citep{veta2019tupac, veta2015assessment, veta2016mitosis}. In contrast, our proposed mitotic score error produced $\overline{ME}$ values of 0.351 and 0.183, respectively, emphasizing its efficacy for mitotic score estimation by showing more differentiation. This difference is further highlighted by comparing three methods (Point-EUNet, Mask-EUNet, and MDFS) on the TUPAC dataset in \cref{fig:mse}b. Despite achieving identical correlation coefficients (0.98), the $\overline{ME}$ metric differentiates the methods' performance with values of 0.15, 0.13, and 0.07, respectively. These results underscore the proposed metric's ability to accurately evaluate mitotic score estimation performance, suggesting its suitability for future comparative studies.

 \begin{figure*}[!ht]
    \centering
    \includegraphics[width=\textwidth]{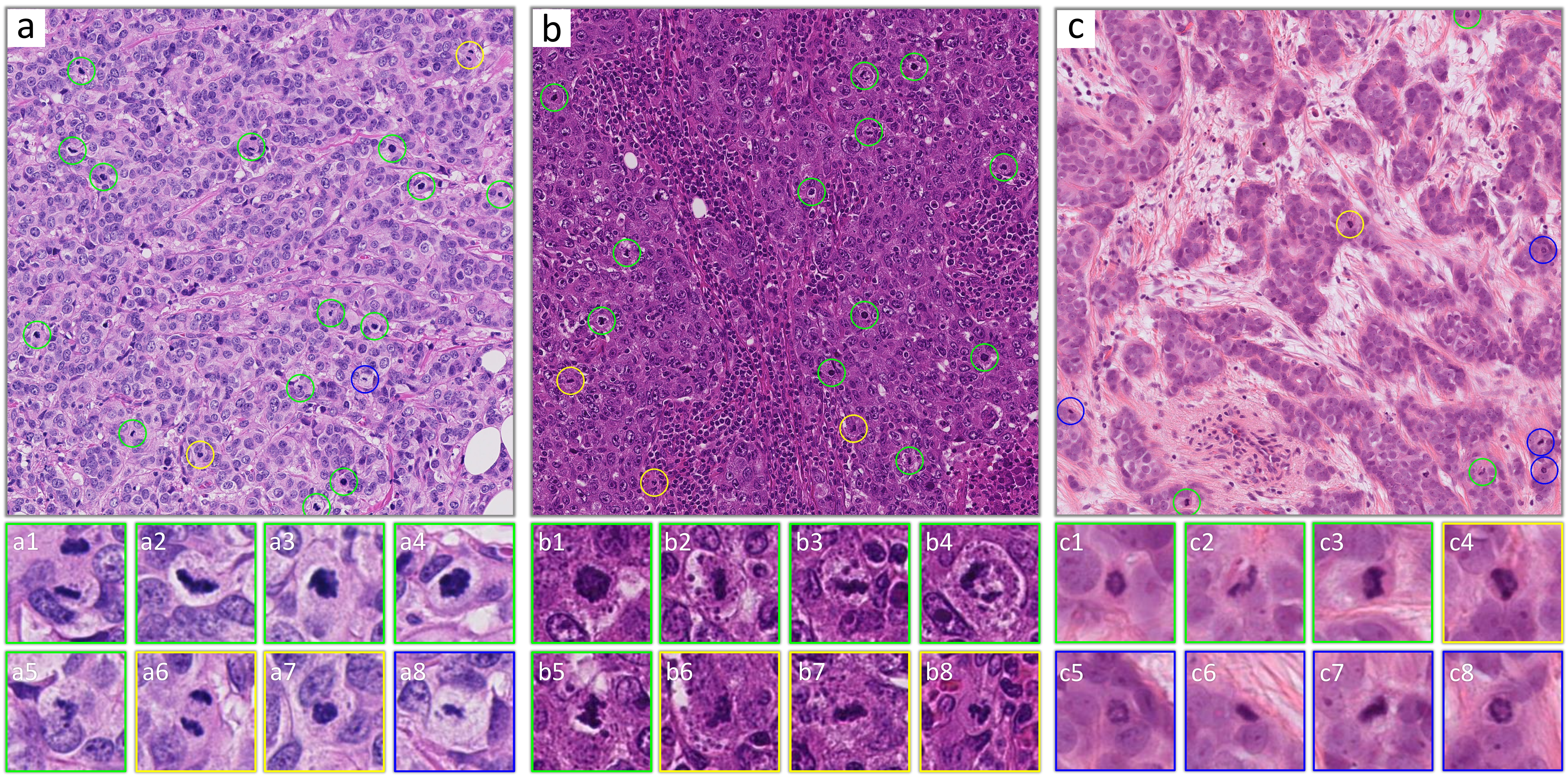}
    \caption{Mitosis detection results of the proposed method on three different images (images in panels `a,c' from MIDOG21, and panel `b' from TUPAC dataset in the top) with the zoomed-in patches of some of the detected mitoses/mimickers in them (panels a\#,b\#, and c\# in the bottom). Circles or patch borders of the color green, blue, or yellow indicate true positive, false negative, and false positive predictions concerning ground truth annotations, respectively.}
    \label{fig:qres}
\end{figure*}

\begin{table*}[!th]
\centering
\caption{Results of external validation experiments on the MIDOG21 test set.}
\begin{tabular}{@{}llccc@{}}
\toprule
\toprule
\multicolumn{1}{c}{Method} & \multicolumn{1}{c}{Detals}                                      & F1                        & Rec                       & Prc                       \\ \midrule
\cite{nateghi2021midog}            & Faster-RCNN + EfficientNet-B0                                   & \multicolumn{1}{l}{0.676} & \multicolumn{1}{l}{0.667} & \multicolumn{1}{l}{0.685} \\
RetinaNet                  & Baseline detection with stain normalization and data aug.       & 0.698                     & 0.699                     & 0.696                     \\
\cite{razavi2021midog}             & Cascade R-CNN                                                   & 0.706                     & 0.706                     & 0.708                     \\
\cite{linag2021midog}             & Fused Detector and Deep Ensemble Classification                 & 0.706                     & 0.686                     & 0.727                     \\
\cite{wilm2021midog}             & RetinaNet with Domain Adversarial branch                        & 0.710                     & 0.731                     & 0.690                     \\
\cite{chung2021midog}             & RetinaNet-101 with Style transfer augmentation                  & 0.724                     & 0.678                     & \textbf{0.776}            \\
\cite{fick2021midog}              & Mask R-CNN + ensemble of ResNet and DenseNet                    & 0.736                     & 0.709                     & 0.764                     \\
\cite{yang2021midog}              & Sk-UNet Model with Fourier Domain for Mitosis Detection         & \textbf{0.747}            & 0.741                     & 0.753                     \\
MDFS (proposed)             & EUNet (Detecting Fast) + EfficientNetB7 (Detecting Slow) & \textbf{0.747}            & \textbf{0.762}            & 0.733                     \\ 
\bottomrule
\bottomrule
\end{tabular}
\label{tab:midog-ext}
\end{table*}

\begin{table*}[!th]
\centering
\caption{Results of external validation experiments on the TUPAC test set. Reported metrics are macro-averages following the TUPAC convention.}
\begin{threeparttable}
\begin{tabular}{lllll}
\toprule\toprule
\multicolumn{1}{c}{Methods}         & \multicolumn{1}{c}{Details}                                     & \multicolumn{1}{c}{F1} & \multicolumn{1}{c}{Rec} & \multicolumn{1}{c}{Prc} \\ \hline
Heidelberg$^\dag$                      &   Combine residual networks with Hough voting and hard-negative mining                                                             & 0.481                  & -                       & -                       \\
\cite{tellez2018whole}                       &   Multi-CNN patch classifiers -- Use private auxiliary data                                                              & 0.541                  & -                       & -                       \\
PIEAS$^\dag$   &  CNN patch classifier using auxiliary ICPR2012/2014 datasets                                                               & 0.571                  & -                       & -                       \\
Microsoft Research$^\dag$             & Shallow CNN patch classifier using auxiliary ICPR2012/2014 datasets                                                                & 0.596                  & -                       & -                       \\
Contextvision$^\dag$                       &   CNN pixel classifier similar to \citep{cirecsan2013mitosis} and hard-negative mining                                                              & 0.616                  & -                       & -                       \\
CUHK$^\dag$ &   Custom CNN patch classifier using auxiliary ICPR2012/2014 datasets                                                              & 0.620                  & -                       & -                       \\
HUST$^\dag$                                &   Custom CNN patch classifier with hard-negative mining                  & 0.626                  & -                       & -                       \\
\cite{akram2018tupac}   & ResNet patch classifier trained on TUPAC+external dataset       & 0.640                  & -                       & -                       \\
IBM Research$^\dag$                        &  Patch classification using Wide residual network and hard negative mining                                                               & 0.648                  & -                       & -                       \\
Lunit$^\dag$                   &  ResNet patch classifier with hard-negative mining                                                               & 0.652                  & -                       & -                       \\
\cite{li2019tupac}         & Weakly (point) supervised mitosis segmentation concentric loss  & 0.669                  & -                       & -                       \\ \hline
EUNet               & Only segmentation part (Detecting Fast)                          & 0.629                  & 0.794                   & 0.521                   \\
MDFS (proposed)                    & EUNet (Detecting Fast) + EfficientNetB7 (Detecting Slow) & 0.675                  & 0.770                   & 0.600                   \\ \bottomrule\bottomrule
\end{tabular}
\begin{tablenotes}
      \item $\dag$\textit{No further information/citation is found for these entries except for the provided details.}
    \end{tablenotes}
\end{threeparttable}
\label{tab:tupac-ex}
\end{table*}

\subsubsection{External validation}
\label{sec:ext-cval}

Our proposed method is tested on 34 TUPAC dataset images and results were submitted for evaluation. The external validation results are detailed in \cref{tab:tupac-ex}, comparing our proposed EUNet model and MDFS pipeline performance against the 11 leading methods in TUPAC's 2016 Task 3 (mitosis detection) challenge. Only three out of these methods have detailed algorithm descriptions (Radboud \citep{tellez2018whole}, Warwick \citep{akram2018tupac}, and SegMitos \citep{li2019tupac}), with the rest summarized in the TUPAC challenge paper \citep{veta2019tupac}. TUPAC challenge participants reported solely macro-average F1 values.


Our method, when evaluated on the TUPAC test set (without external data), outperforms all other methods (achieving a macro-average F1 of 0.675). This underscores the efficacy of our `Detecting Slow' system where the macro F1 improves by 5\% primarily by boosting precision upon the addition of a deep classifier atop our EUNet candidate segmentation model. This enhancement is made at a slight computational speed sacrifice (\cref{tab:downsample}, \cref{sec:int-cval}). Hence, our method outperforms the challenge winner, Lunit, which managed a macro F1 of 0.652. Subsequent post-challenge submissions, specifically the SegMitos method \citep{li2019tupac}, yield comparable results to ours (F1=0.669). Nevertheless, our method significantly surpasses all other techniques, which typically merge patch classifiers with hard-negative mining.


The MIDOG test set comprises 80 images from diverse breast tumor instances scanned using four different scanners, two of which were utilized for the MIDOG training set acquisition. Methods are evaluated by submission to the challenge platform, similar to TUPAC. Our method's results, alongside the top eight performing methods in the MIDOG21 challenge's final testing, are displayed in \cref{tab:midog-ext}. Our algorithm ranks top (tied with \cite{yang2021midog}), winning the challenge with an F1 of 0.747 and the highest recall of 0.762 (precision of 0.733). It proves superior to all region proposal-based algorithms such as bounding box detection algorithms \citep{nateghi2021midog,lin2017retinanet,razavi2021midog,linag2021midog, wilm2021midog,chung2021midog} and Mask-RCNNs \citep{fick2021midog}, raising the detection F1 score by roughly 5\% compared to the RetinaNet baseline.

\begin{figure}[!t]
    \centering
    \includegraphics[width=\columnwidth]{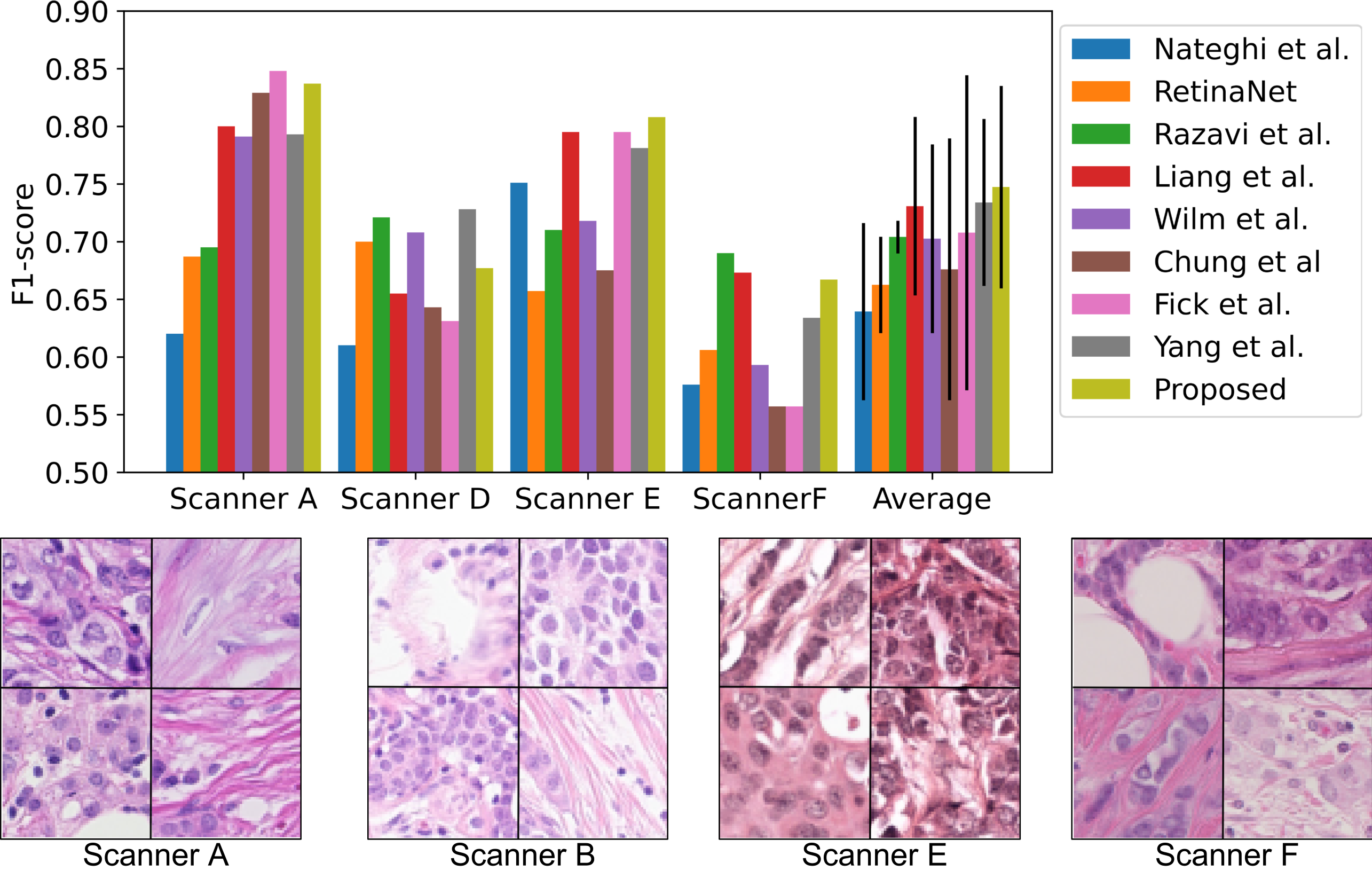}
    \caption{Results of the top performing methods of the MIDOG21 challenge on images from different scanners.}
    \label{fig:scanner}
\end{figure}

The bar chart of F1-scores of top methods over the cases from the four scanners in the test set is presented in \cref{fig:scanner} to examine the method's performance across different sources. Only images from `Scanner A' are used in training, with the other three classified as `out-of-domain' scanners. Our method consistently performs well across all scanners, especially `Scanner A' and `Scanner E' (achieving F1 of 0.837 and 0.808 respectively). With an F1 of 0.677, `Scanner D' results fall outside the top 3 performers. Conversely, \cite{yang2021midog} achieved F1 of 0.726 potentially due to their unique Fourier domain-mixing algorithm, incorporating unlabelled `Scanner D' images into training. Finally, we computed the macro-average F1-score by independently averaging the F1-score for each scanner. This yielded a superior macro-average F1 of 0.747$\pm$0.08 across all scanners, compared to \citep{yang2021midog} (0.734$\pm$0.07) and other methods.

\subsubsection{Independent cross-validation}

Finally, we aim to demonstrate the generalizability of the proposed model on external data from different dataset sources. To this end, we train our MDFS method on either TUPAC or MIDOG21 data, before testing it on the training set of the other dataset. In addition, we report the results of TUPAC and MIDOD21 models on the ICPR dataset. The images of the test sets in this experiment are not only from different sources but also have been annotated with different annotation protocols.
The results for these experiments are reported in \cref{tab:indep_cval}. We observe that the model trained on the TUPAC data shows good generalizability to other datasets, achieving F1 scores in the range of 0.758 and 0.745 on the MIDOG training set and ICPR2012 test set, respectively. Similarly, the model trained on the MIDOG dataset also demonstrates competitive performance with F1 scores, especially on the ICPR2012 test set. It is important to note that due to variations in annotation protocols (posterior shift) and potential prior shifts (variations in the distribution of labels in different datasets), we refrain from making direct comparisons regarding the superiority of one model over the other.

\begin{table}[h]
\centering
\caption{Independent cross-validation experiments where the model is trained on the `Source Dataset` and tested on the `Target Dataset'.}
\begin{tabular}{@{}llccc@{}}
\toprule
\toprule
\textbf{Source dataset} & \textbf{Target dataset} & \textbf{F1} & \textbf{Rec} & \textbf{Prc} \\ \midrule
TUPAC          & MIDOG21           & 0.758       & 0.714           & 0.807        \\
MIDOG21           & TUPAC           & 0.697       & 0.708           & 0.686        \\
MIDOG21          & ICPR2012             & 0.736       & 0.685           & 0.796        \\
TUPAC           & ICPR2012           & 0.745       & 0.695           & 0.803        \\
\bottomrule \bottomrule
\end{tabular}
\label{tab:indep_cval}
\end{table}

\subsection{Added value of `Detecting Fast' and `Detecting Slow'}
\label{sec:time}

The `Detecting Fast' system's efficacy in our MDFS framework was assessed by testing various down-sampling scales for the resizing module preceding the mitosis candidate segmentor (\cref{fig:overview}b). Stain-normalized images were used without stain augmentation for these experiments to prevent potential random effects due to data variability, ensuring fair comparisons with RetinaNet and EfficientDet. Consider the 0.75 down-scaling ratio from \cref{tab:downsample} for an example, although performance metrics seem reduced (F1=0.740) compared to using full image resolution (F1=0.754), the F1 is only 0.004 lower after applying `Detecting Slow' (i.e., MDFS pipeline) (0.785 for full resolution and 0.781 when down-scaled by 0.75), while being about 1.8 times faster. We use the EUNet algorithm at full resolution (scale=1) as a time baseline and indicate other algorithms' speed gain relative to it in \cref{tab:downsample}. The proposed EUNet and full pipeline clearly outpace EfficientDet and RetinaNet bounding box detection models, although deeper encoders may provide better F1 scores for bounding box detection models at increased computation time.


 Moreover, comparing EUNet's running time and speed gain with the full MDFS pipeline at each scale reveals that adding `Detecting Slow' atop `Detecting Fast' incurs minimal computational overhead (for instance, running time only reduces about 50 ms for scale 1, resulting in 0.98$\times$ speed gain) while significantly enhancing detection F1 (about 5\% for scale 0.5). This is primarily because the proposed pipeline's second system only processes a small number of mitosis candidates, which are smaller patches and take less time. Benchmarking experiments were run on a Nvidia DGX-2 device with one Tesla V100 GPU.

\subsection{Qualitative assessment}
\label{sec:qres}
\begin{figure*}[!ht]
    \centering
    \includegraphics[width=1.0\textwidth]{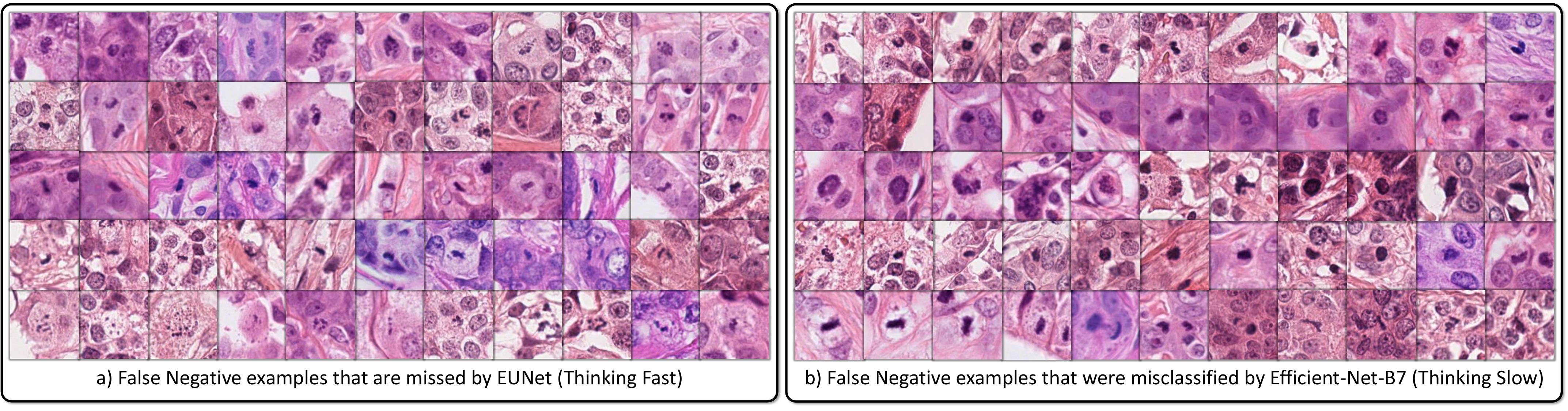}
    \caption{Challenging mitotic figures that were missed by MDFS framework either in candidate segmentation phase (a) or candidate refinement phase (b). Images are collected from the results on MIDOG21 dataset.}
    \label{fig:fn}
\end{figure*}


In our study, we also conduct a qualitative evaluation of our mitosis detection approach. We randomly select three images, images a and c from the MIDOG dataset and image b from the TUPAC dataset, for in-depth analysis (\cref{fig:qres}). In these larger images, TP, FP, and FN detections are denoted by green, yellow, and blue circles, respectively. We pick eight unique detections from each image for detailed visualization. We randomly choose TPs and display all FPs and FNs beneath the associated larger image, color-coding the detection boundaries to match the circles above.

Our examination of \cref{fig:qres} revealed inconsistencies in mitotic figure annotations in both TUPAC and MIDOG21 datasets. We confirmed these inconsistencies by consulting pathologists who evaluated the selected detections (highlighted below the larger images in \cref{fig:qres}). Though their expert opinion was sought, we acknowledge mitosis interpretation is not definitive and can vary among pathologists \citep{veta2016mitosis, ibrahim2022assessment,alkhasawneh2015interobserver}. We mitigated bias by blinding the pathologists to the patches' association with our algorithm's detections and providing full images for context.
The pathologists affirmed that all patches, excluding c6, contained mitotic figures. This indicates that the original annotators may have overlooked six mitoses in these ROIs. Interestingly, our pathologists identified an additional mitotic figure missed by both the original annotators and our algorithm. However, their validation was limited to typical examples that are usually recognizable by breast pathologists.

Our observations corroborate existing literature on the difficulties of mitotic figure detection and inter-observer variability implications \citep{veta2016mitosis, ibrahim2022assessment,alkhasawneh2015interobserver,saldanha2020global,bertram2020pathologist,molenaar2000observer,lashen2021visual,ibrahim2023improving}.They also underscore the need for continuous refinement of annotation process (by using PHH3 IHC staining \citep{alkhasawneh2015interobserver,ibrahim2023improving,tellez2018whole}, using well-established annotations protocols/guidelines \citep{lashen2021visual,ibrahim2022assessment}, or consensus of pathologists observations \citep{wilm2021influence,bertram2019large}) and detection algorithms to account for the inherent limitations and complexities of datasets. This nuanced understanding informs the development and evaluation of mitotic detection methodologies, emphasizing both the strengths and potential areas of improvement.


Despite the decrease in the false positive rate for the EUNet segmentation model after the hard-negative mining phase (see \cref{tab:downsample,tab:tupac-cval} where MDFS precision increases over EUNet), there remains a risk of discarding true positives during classification.
In \cref{fig:fn}, we present 110 false negative figures overlooked by EUNet in the `Detecting Fast' system or misclassified by the EfficientNet-B7 model in the `Detecting Slow' system. Most missed samples in \cref{fig:fn} are small, particularly those overlooked in the first phase (fourth row of \cref{fig:fn}a). Also, anaphase mitotic figures with distantly spaced daughter cells pose challenges for both segmentation and classification tasks (second row of \cref{fig:fn}a). Some mitotic figures resemble inflammatory cells and are thus misclassified in the second phase (first row of \cref{fig:fn}b), and some atypical mitotic figures are even difficult for the `Detecting Slow' system (third row of \cref{fig:fn}b).

\subsection{Ablation studies}
\label{sec:ablation}

To optimize our framework and assess its different aspects, we conducted ablation studies using the MIDOG21 training set and 3-fold cross-validation as detailed in \cref{sec:midog}.
In these experiments, the scaling factor in the `Detection Fast' system was set to 1 to fully utilize the data.

\subsubsection{Supervision for EUNet}
\label{sec:supervision}


We trained our EUNet alongside RetinaNet \citep{lin2017retinanet} and EfficientDet \citep{tan2020efficientdet} on candidate detection, using dilated point annotations as ground truth masks (Point-EUNet). It achieved F1 score, recall, and precision of 0.731, 0.775, and 0.693, respectively. This lets us compare supervisory signal performance. The EUNet trained on mitosis masks (Mask-EUNet) attained an F1 of 0.754, significantly outperforming both bounding box detection methods (see \cref{tab:downsample}). Intriguingly, Point-EUNet outperformed both bounding box models but was still inferior to Mask-EUNet. We propose that using a segmentation model for mitosis candidate detection is a suitable approach.


\subsubsection{Effect of domain generalization techniques}

We examine three different techniques to tackle the domain-shift issue in histology images: stain normalization (SN), stain augmentation (SA), and encoder pre-training (see \cref{sec:domain-shift}).
Firstly, we test our candidate segmentation model pretrained with various self-supervised learning methods on the MIDOG21 dataset. The methods include \textit{ImageNet}, SimCLR \citep{chen2020simclr}, and SCL \citep{khosla2020scl} for the model encoder, and the proposed SSHL method for both the encoder and decoder. No other domain generalization techniques are used here to ensure optimal pretraining method selection. As shown in \cref{tab:pretrain}, SSHL notably outperforms both SimCLR and SCL pretraining algorithms by 2.5\%, achieving an F1 of 0.741. Thus, we only consider the SSHL pretraining method in subsequent domain generalization investigation experiments (ImageNet weights are also included as the standard approach).

\begin{table}[!h]
\centering
\setlength{\tabcolsep}{5pt}
\caption{Effect of using different pretraining methods on the performance of the mitosis candidate segmentation model.}
\begin{tabular}{llccc} 
\toprule\toprule
\textbf{Method}     & \textbf{Pretrain}  & \textbf{F1}    & \textbf{Rec}   & \textbf{Prc}    \\ 
\hline
ImageNet             & Encoder      & 0.722          & 0.760          & 0.687           \\
SimCLR               & Encoder     & 0.716          & 0.735          & 0.698           \\
SCL                  & Encoder     & 0.716          & 0.703          & \textbf{0.729}           \\
Proposed-SSHL        & Encoder-Decoder     & \textbf{0.741}          & \textbf{0.807}          & 0.685           \\
\bottomrule\bottomrule
\end{tabular}
\label{tab:pretrain}
\end{table}


In \cref{tab:domain-shift}, we present the effect of SN, SA, and pretraining techniques on the mitosis candidate segmentation task (Detecting Fast system). As expected, without using any of the proposed techniques, the model performs the worst (Ex.1 with F1 of 0.722). With ImageNet pretrained weights, the model performance consistently improved with the addition of SN and SA. A similar pattern was observed with SSHL pretraining, though SN and SA had less impact on the final F1. Interestingly, when combined with SN and SA, ImageNet pretrained weights outperformed the SSHL pre-trained model (Ex.4 with F1 of 0.773). These results suggest that SSL techniques are beneficial for introducing domain-invariance to mitotic segmentation models. However, when combined with SN and SA techniques, they might be unnecessary. Thus, we used ImageNet weights for encoder pretraining and a combination of SN and SA for training and inference on benchmark datasets. Note that SN was excluded during WSI inference to reduce computational load.

\begin{table}[!h]
\centering
\caption{Effect of using different combinations of domain generalization techniques on the mitosis candidate segmentation model's performance.}
\begin{tabular}{lcccccc} 
\toprule\toprule
\textbf{Ex.} & \textbf{Pretraining}    & \textbf{SN} & \textbf{SA} & \textbf{F1}    & \textbf{Rec}   & \textbf{Prc}    \\ 
\hline
1            & ImageNet        & $\times$       & $\times$       & 0.722          & 0.760          & 0.687           \\
2            & ImageNet        & $\times$       & $\checkmark$   & 0.765          & 0.807          & 0.728           \\
3            & ImageNet        & $\checkmark$   & $\times$       & 0.754          & \textbf{0.824} & 0.695           \\
4            & ImageNet        & $\checkmark$   & $\checkmark$   & \textbf{0.773} & 0.805          & \textbf{0.744}  \\ 
\hline
5            & SSHL & $\times$       & $\times$       & 0.741          & 0.807          & 0.685             \\
6            & SSHL & $\times$       & $\checkmark$   & 0.758            & 0.788            & 0.730             \\
7            & SSHL & $\checkmark$   & $\times$       & 0.748            & 0.775            & 0.724             \\
8            & SSHL & $\checkmark$   & $\checkmark$   & 0.762            & 0.805            & 0.724             \\
\bottomrule \bottomrule
\end{tabular}
\label{tab:domain-shift}
\end{table}

\subsection{Adaptation to other tissues and species}
\label{sec:midog22}

The diversity of tissue types and species in the MIDOG22 dataset provides an ideal test bed for the generalization capability of our MDFS algorithm. However, the MIDOG22 challenge's main track prohibits the use of external resources, precluding the integration of NuClick for mitosis mask generation as discussed in \cref{sec:nuclick}. Instead, we employ Point-EUNet (utilizing dilated points as mitosis GT, see \cref{sec:supervision}) within the `Detecting Fast' system, which can be trained using the original point annotation. Furthermore, unlike the MIDOG21 challenge, we opted not to use stain normalization in MIDOG22.
We conducted cross-validation experiments on the MIDOG22 training set, and to evaluate MDFS on the external MIDOG22 test set, we submitted our algorithm to the challenge. \Cref{tab:midog22} reports our cross-validation results and the top-performing methods on the MIDOG22 test set. Remarkably, MDFS secured the 1\textsuperscript{st} rank on the final test leaderboard\footnote{MIDOG 2022 challenge leaderboard is accessible at \url{https://midog2022.grand-challenge.org/}}.


 \begin{table}[!t]
    \centering
    \caption{Cross-validation and external test results for mitosis detection on MIDOG22 dataset.}
    \begin{tabular}{@{}llccc@{}}
    \toprule\toprule
    Method     & Set                                       & F1 & Rec & Prc \\ \hline
    \cite{yang2021midog}     &Test    & 0.658                  & 0.528                   & \textbf{0.875}                   \\
    \cite{wilm2021midog} &Test & 0.713                  & 0.663                   & 0.771                   \\
    \cite{kotte2022midog22} &Test            & 0.751                  & \textbf{0.764}                   & 0.738                   \\
    \cite{saipradeep2022midog22}  &Test    & 0.756                  & 0.754                   & 0.759                   \\
    MDFS (proposed)    &Test                                                   & \textbf{0.764}                  & 0.712                   & 0.823                   \\ \hline
    MDFS (proposed)    &Train                                                   & 0.816                 & 0.831                   & 0.801                   \\
    \bottomrule\bottomrule
    \end{tabular}
\label{tab:midog22}
\end{table}


Considering that the MIDOG22 test set comprises 100 cases from 10 different unseen tumor types (human melanoma, human astrocytoma, human bladder carcinoma, canine breast cancer, canine cutaneous mast cell tumor, human meningioma, human colon carcinoma, canine hemangiosarcoma, feline soft tissue sarcoma, and feline lymphoma), the high F1 score of 0.764 achieved by the MDFS method demonstrates its robustness and adaptability to domain shifts caused by varying scanners, labs, species, and tumor types. Notably, MDFS outperformed SOTA attention-based transformer models \citep{kotte2022midog22,saipradeep2022midog22}, reaffirming the superiority of our proposed method based on mitosis segmentation over standard bounding box detection models for mitosis detection.

\subsection{Large-scale mitosis detection on TCGA WSIs}

To showcase the capability of our proposed method for the efficient processing of WSIs, we processed the entire breast cohort of the TCGA dataset (TCGA-BRCA) with an improved version of MIDOG22 model (see \ref{sec:tcga-brca}). Over 620K mitotic figures were detected in 1125 WSIs, with the candidate segmentation and refinement parts of the algorithm requiring around 2.5 ($\pm$6) minutes and 6.4 ($\pm$4.5) seconds, respectively. Thus, each slide was processed in under 3 minutes. Given its high efficiency in processing large WSIs and its robustness to scanner-induced variations, our algorithm is a viable tool for research uses that demand WSI-level mitosis detection.


We have made the output of our mitosis detection algorithm publicly accessible for research purposes at \url{https://sandbox.zenodo.org/record/1227403}\footnote{This link contains only 10 samples of TCGA-BRCA and the full TCGA-BRCA-Mitosis dataset will be released when the journal version of the manuscript is published.} to facilitate mitosis-related down-stream tasks, such as biomarker discovery and survival prediction for breast cancer. The `TCGA-BRCA Mitosis Dataset' also includes mitotic hotspot regions, hotspot mitotic counts, and hotspot mitotic scores. For more details, refer to \ref{sec:tcga-brca}.

\section{Discussion}
\label{sec:discussion}

Various algorithms for automatic mitosis detection have been proposed \citep{mathew2021survey}, with many aimed at enhancing detection in mitotic hotspots or regions of interest (ROIs). This paper outlines an efficient, generalizable algorithm for mitosis detection, designed to be resilient to domain shifts caused by scanner variability, cancer types, or species. This two-stage algorithm initially segments lower-resolution mitosis candidates and then refines them at a higher resolution for improved speed and accuracy. We introduce a new metric, mitotic score error $\overline{ME}$, to better assess the performance of mitotic score estimation methods.

\subsection{Generalizability of MDFS}

Our proposed method's robustness is evaluated through external validation experiments (\cref{sec:int-cval,sec:ext-cval,sec:midog22}). These trials highlight the MDFS's adaptability to unseen domains, including those arising from variations in staining, scanner use, tissue types, case species, or annotation protocols. Our method achieved first place in both the MIDOG2021 and MIDOG22 grand challenges on mitosis domain generalization and outperformed all other techniques on the TUPAC-mitosis challenge leaderboard.


Notably, the MIDOG21 dataset's test set comprises images from four different scanners. Our algorithm achieves an F1 of 0.837 when tested on the `Scanner A' subset (\cref{fig:scanner})—a 5\% improvement over internal cross-validation experiments. We attribute this high generalizability to the use of domain generalization techniques (\cref{sec:domain-shift}) and model design. Future research should continue utilizing encoder-decoder models and stain-augmentation techniques to combat domain shifts in mitosis detection. While we didn't observe any added value in pretraining the segmentation model for our current work, it could potentially prove beneficial when handling small-scale datasets. Furthermore, despite potential advantages, we caution against the use of stain normalization due to potential inconsistencies, particularly when original images contain staining components other than Hematoxylin and Eosin \citep{dang2022nuclear}.

\subsection{Efficiency of MDFS}

 The proposed algorithm outperforms other region-proposal-based techniques in processing $2mm^2$ sample images (\cref{sec:int-cval}). This efficiency owes to the method's dual detection systems. The `Detecting Fast' system quickly identifies potential mitotic figures in down-scaled images, while the `Detecting Slow' system accurately classifies candidates using full-resolution patches. Despite the initial loss of high-resolution information due to down-sampling, our two-step approach regains much of the lost performance (\cref{tab:downsample}). The 'Detecting Slow' system can improve the F1 score by 3\%-5\%, while minimally impacting computational cost (\cref{sec:time}). 
 

As seen in \cref{tab:downsample}, the proposed method maintains mitosis detection accuracy using down-scaled images in the `Detecting Fast' system, while significantly enhancing speed. With a down-sampling scale of 0.75, the 'Detecting Fast' system sustains an F1 value above 0.78, reducing $2mm^2$ ROI processing time from 2.8 seconds to 1.5 seconds, an approximate 86\% speed improvement. On average, WSIs are processed in about 3 minutes. This demonstrates the MDFS system's practical efficiency and the generalizability of EUNet with lower-magnification images. Contrarily, methods such as EfficientDet markedly deteriorate in performance at lower resolutions and show sensitivity to threshold selection, rendering them ill-suited for employment in the `Detecting Fast' system.

\subsection{Limitations and future work}
As mentioned in \cref{sec:qres}, one of the main drawbacks of the MDFS method is missing very small/faint or inflammatory-like mitoses. It appears, however, many of the false negatives (FNs) in \cref{fig:fn} are mislabeled mimickers, not actual mitoses. For example, dissolved nuclear materials in the samples of bottom row in \cref{fig:fn}a exhibit traits of dead or dying cells, which are called Karyorrhectic cells \citep{ibrahim2022assessment}. In general, \cite{ibrahim2022assessment} proposed guidelines on recognizing common mimickers in breast cancer listed as apoptotic bodies, tissue artifact (pigmentation), hyperchromatic malignant cells, foamy macrophages, karyorrhectic cells, and out-of-focus lymphocytes/fibroblast cells which are challenging for both pathologists and AI to be distinguished.


An additional issue is the considerable number of true positive (TP) candidates identified by the `Detecting Fast' system pruned during refinement. Thus, a highly sensitive and specific classifier is necessary. Ensembling multiple classifiers has shown efficacy in enhancing classifier performance \citep{linag2021midog,kotte2022midog22}, although at the cost of computational power and decreased algorithmic speed. Essentially, for optimal functionality of `Detecting Fast and Slow', both segmentation and classification components must perform exceptionally.


A crucial facet of automated mitosis detection is determining its potential to enhance survival prediction, as patient prognosis is ultimately the aim of mitotic counting in slides. Given that our algorithm can perform on par with experienced pathologists on mitosis detection in ROIs \citep{aubreville2022midog}, AI-assisted mitotic scoring is anticipated to improve survival prediction accuracy. Nonetheless, this topic and evaluation of mitosis detection accuracy on the WSI level have not been probed yet and exceed the current study's scope.
Further, it would be insightful to scrutinize the performance of our proposed mitosis detection algorithm across other cancer types. Automation of mitosis detection in such instances could contribute to developing objective prognostic measures for patients.

\section{Conclusions}

This paper presented a two-stage algorithm for effective mitosis detection in breast histology images and WSIs, involving an initial `Detecting Fast' phase to segment mitotic candidates, followed by a `Detecting Slow' phase to refine these candidates using a deeper CNN. We introduced the EUNet model for mitosis segmentation and utilized Efficient-Net \citep{tan2019efficientnet} for candidate classification. We demonstrated that the 'Detecting Fast' phase could employ lower-resolution images to substantially boost algorithm speed without compromising accuracy. Our investigation into the effect of three domain generalization techniques on the mitosis detection task indicated that a combination of stain normalization and augmentation techniques yielded optimal results. Self-supervised pretraining of the encoder model, even with a novel preprocessing method capable of joint encoder and decoder pretraining, was found to be unnecessary with our method and mid- to large-scale datasets. Our approach outperformed all other SOTA methods for mitosis detection on MIDOG and TUPAC datasets, known for significant domain shifts. This performance advantage is evident in terms of both traditional detection metrics and the recently proposed mitosis score error, $\overline {ME}$, which assesses the mitosis detection model's performance on mitosis score estimation in hotspot regions.


Our algorithm secured first places in the MIDOG 2021 and 2022 mitosis detection challenges and outperformed all other methods on the TUPAC dataset. Furthermore, we processed 1125 WSIs of the TCGA-BRCA cohort using our efficient method and generated over 530K mitotic figures. This dataset, along with the mitosis masks produced for the TUPAC and MIDOG datasets, is made publicly available to aid in the development of mitosis detection models and mitosis-based survival analysis for breast cancer.

\section*{Acknowledgments}
The authors would like to thank the collaborating pathologist Asmaa Ibrahim, MD, and Ayat Lishen, MD, from Nottingham Breast Cancer Research Center for cross-checking the FPs and FNs reported in \cref{fig:qres}. MJ, SG, SEAR, FM, and NR report financial support provided by UK Research and Innovation (UKRI). NR and FM report financial support from GlaxoSmithKline, outside of the submitted work. AS reports financial support from Cancer Research UK (CRUK), outside of the submitted work. NR and SG are co-founders of Histofy Ltd.

\appendix
\setcounter{figure}{0}
\renewcommand{\thefigure}{A\arabic{figure}}

\section{Threshold selection}
\label{sec:thresh}
\begin{figure*}[!ht]
    \centering
    \includegraphics[width=0.85\textwidth]{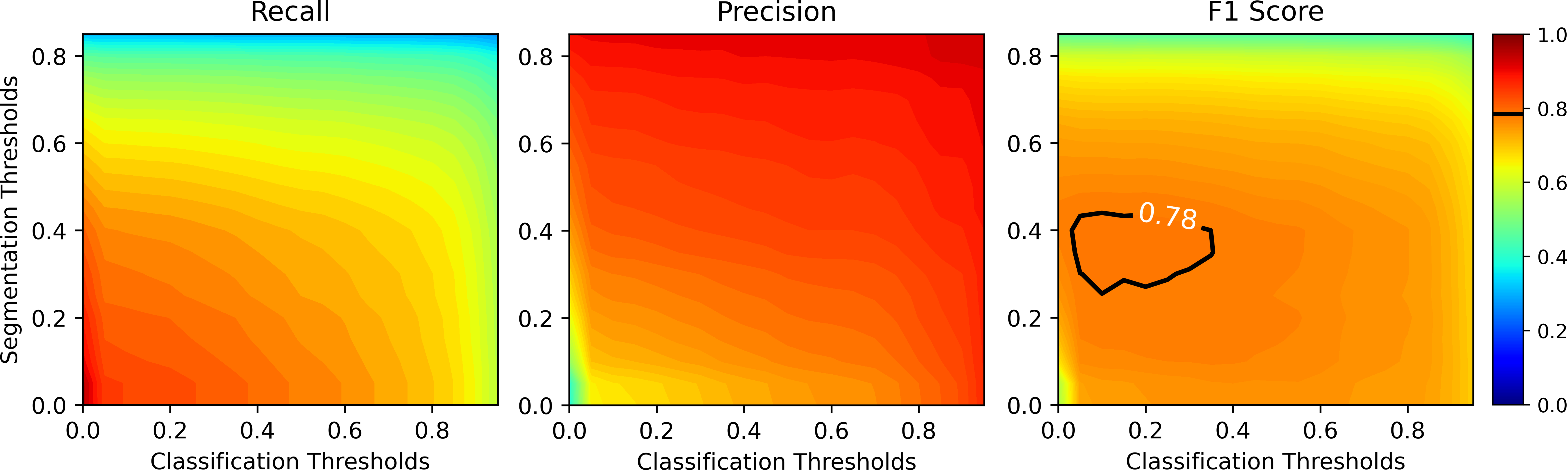}
    \caption{Threshold analysis experiment on MIDOG training set: Recall, Precision, and F1 values are highlighted against various selections of segmentation and classification thresholds during the post-processing step. Black contour shows peak performance region.}
    \label{fig:thresh}
\end{figure*}

The post-processing design is impactful on the final output and should be carefully designed. The simple segmentation and classification design of our model allows the incorporation of a fast post-processing method (\cref{sec:postproc}). There are two main parameters in the proposed post-processing and the entire prediction pipeline, which are segmentation and classification thresholds. We set these parameters by conducting a grid search on a range of thresholds for both tasks. In particular, we sweep the segmentation threshold between 0 to 0.85 and for each segmentation threshold, we test a range of classification thresholds from 0 to 0.95 (both inclusive). For each threshold configuration, we calculate the F1, Recall, and Precision of the results against the GT and plotted the results as surface maps in \cref{fig:thresh}. The best-performing point is chosen based on the F1 map, where a good combination of recall and precision leads to high values of the F1.

Through careful examination of the heatmaps \cref{fig:thresh}, one can observe the complex trade-offs between sensitivity and specificity in the initial stage of our two-stage detection process. High detection thresholds in the first stage yield higher specificity but lower sensitivity, resulting in fewer but more accurate candidate detections. However, this stringent selection might miss some true positive cases. By contrast, lower detection thresholds increase sensitivity, at the risk of passing more false positives to the subsequent classification stage. To compensate for the lower specificity in the first stage, the second stage employs a range of classification thresholds. With the help of the classification heatmap, it is possible to select an optimal classification threshold that best balances precision and recall, effectively refining the set of candidates passed on from the first stage. This allows us to maintain an excellent F1 score, despite the trade-offs made in the initial stage. Therefore, by appropriately tuning the thresholds in both stages, we can achieve efficient and robust performance in mitotic figure detection.

A benefit of our proposed method is that resulted F1 map is almost flat for all spans of segmentation and classification thresholds and this shows the robustness of the proposed method against threshold selection. In particular, the peak performance (where the F1 is higher than 0.78 and highlighted by a black contour in \cref{fig:thresh}) can be achieved by selecting the segmentation threshold in the range of $[0.3, 0.45]$ and the classification threshold anywhere between $[0.1, 0.4]$ which is indicative of the robustness of our method against threshold selection. That said, our approach does allow for flexibility. If a user seeks to optimize for sensitivity (at the cost of precision), they can select lower thresholds. Conversely, if precision is the priority (at the cost of sensitivity), higher thresholds can be selected. Our method's distinctive dual advantage of robustness and flexibility in threshold selection, demonstrated by consistent performance across varying thresholds yet allowing adjustments for a desired balance of sensitivity and precision, underscores its potential to be a practical tool for mitotic figure detection.
Please note that this figure is related to the experiment where we used stain-normalized images and train our models with the stain-augmentation technique.
For other experiments, these maps might be slightly different but it has been seen that the landscape of F1 values in relation to segmentation and classification thresholds is almost always flat with different variations of our model. Nonetheless, for each cross-validation experiment, we repeated a similar threshold selection procedure to select the segmentation and classification thresholds that best suit the configuration of the method to make fair comparisons in the paper.

\begin{figure}[!ht]
    \centering
    \includegraphics[width=\columnwidth]{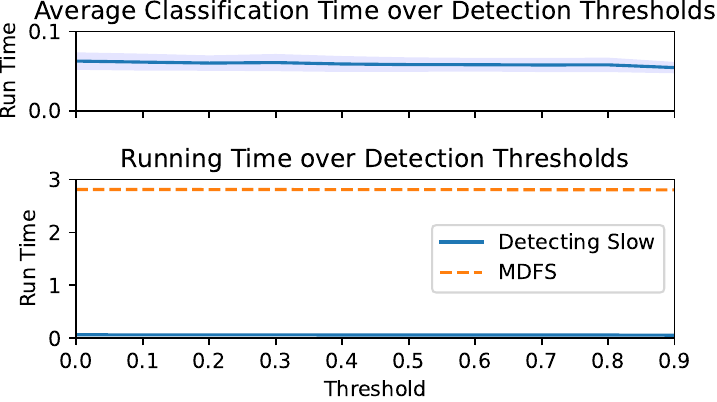}
    \caption{Analysis of `Detecting Slow' and `MDFS' run times (in seconds) over varying detection thresholds. The upper plot shows the average classification time for `Detecting Slow', with shading indicating standard deviation. The lower plot juxtaposes running times for both systems. The x-axis is common and represents thresholds used in the `Detecting Fast' system.}
    \label{fig:timing}
\end{figure}

In addition, to address concerns about computational efficiency, we have plotted the runtime impact of varying detection thresholds on the classification stage and the whole MDFS piple in \cref{fig:timing}. Our findings suggest that while the detection threshold significantly influences the number of candidates presented to the classification stage, the added computational load does not substantially affect overall runtime. This observation underscores the efficiency of our two-stage approach, even when adjusting for higher sensitivity in the initial detection phase.

\section{Details of TCGA-BRCA mitosis dataset}
\label{sec:tcga-brca}

\begin{figure*}[!th]
    \centering
    \includegraphics[width=\textwidth]{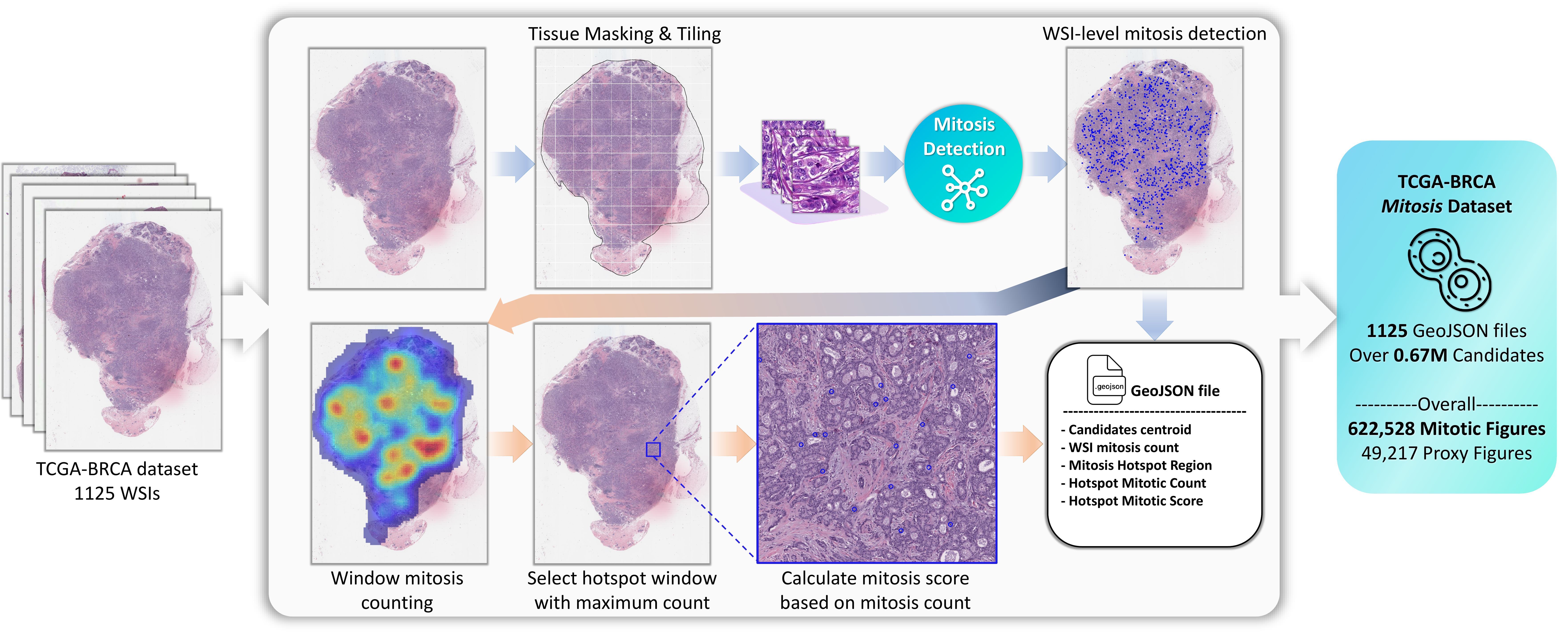}
    \caption{Mitosis detection pipeline in WSIs and TCGA-BRCA mitosis dataset generation. Mitotic figures are first detected in each WSI and then based on mitosis density a hotspot region is extracted to calculate mitotic count and mitotic score.}
    \label{fig:tcga}
\end{figure*}

The WSI processing pipeline is shown in \cref{fig:tcga} and explained in \cref{sec:method-WSI}. In our processing of the TCGA-BRCA dataset, we relied on the TIAToolbox software, which effectively manages image scaling. We extracted tiles from WSIs at a resolution of 0.25 mpp, equivalent to roughly a 40x magnification, then feed them into the MDFS pipeline which down-scales the image tile by a factor of 0.75 in the `Detecting Fast' system. When encountering WSIs scanned at higher magnifications, such as 80x, TIAToolbox performed the necessary downscaling of tiles to match our desired resolution. As for the 20x slides, we ensured high-quality detection by utilizing segmentation and classification models trained on half-scaled (20$\times$ images).

In order to deal with large variability of mitoses and usual artifacts in WSIs (such as pen markings and stain residues), we fine-tuned the MIDOG22 classifier model on a manually curated dataset of common artifacts in histopathology. Small (128$\times$128 pixels) artifact patches were extracted from a selection of TCGA and an in-house dataset to form a collection of 22,600 artifact images. Doing this has made our `Detecting Slow' system more robust against obvious artifacts in WSIs.

In total, our released `TCGA-BRCA Mitosis Dataset' comprises 1125 annotation files in JSON format containing more than 0.67 million candidates (initially detected by the segmentation model), of which 622,528 are confirmed to be mitoses by the `Detecting Slow' classifier. We have released both mitosis figures and proxy figures (instances that were pruned out by the `Detecting Slow' system but had a mid-range probability of being mitosis) to further aid in developing better mitosis detection models and downstream analysis in the future. For each WSI in the dataset, we release the candidates' centroid, bounding box, hotspot location, hotspot mitotic count, and hotspot mitotic score. This dataset can be found at  \url{https://sandbox.zenodo.org/record/1227403}.
It should be noted that we did not conduct a comprehensive review of all mitotic figures within each WSI, and we do not purport these to be free of errors. Nonetheless, a team of two pathologists examined the resultant hotspot regions of interest from over 700 WSIs within the TCGA-BRCA-Mitosis dataset. This examination aimed to verify the quality of the selections, ensuring they were not primarily driven by excessive false detections or artifacts.

\bibliographystyle{model2-names.bst}
\biboptions{authoryear}
\bibliography{refs}




\end{document}